\pdfoutput=1
\documentclass[journal,twoside,web]{ieeecolor}
\usepackage{tmi}
\usepackage{cite}
\usepackage{amsmath,amssymb,amsfonts}
\usepackage{algorithmic}
\usepackage{graphicx}
\usepackage{textcomp}
\usepackage{etoolbox}
\makeatletter
\@ifundefined{color@begingroup}%
  {\let\color@begingroup\relax
   \let\color@endgroup\relax}{}%
\def\fix@ieeecolor@hbox#1{%
  \hbox{\color@begingroup#1\color@endgroup}}
\patchcmd\@makecaption{\hbox}{\fix@ieeecolor@hbox}{}{\FAILED}
\patchcmd\@makecaption{\hbox}{\fix@ieeecolor@hbox}{}{\FAILED}
\usepackage{graphicx}
\usepackage{amsmath}
\usepackage{amssymb}
\usepackage{booktabs}
\usepackage{xspace}
\usepackage{multicol}
\usepackage{cuted}
\usepackage{etoolbox}
\usepackage{float}
\usepackage{multirow}
\usepackage{makecell}
\usepackage{hyperref}
\hypersetup{
  colorlinks=true,
  linkcolor=blue, 
}
\usepackage{tabu}
\usepackage{pifont}
\usepackage{colortbl}
\usepackage{adjustbox}
\makeatletter
\DeclareRobustCommand\onedot{\futurelet\@let@token\@onedot}
\def\@onedot{\ifx\@let@token.\else.\null\fi\xspace}

\def\eg{\emph{e.g}\onedot} 
\def\ie{\emph{i.e}\onedot} 
 
\def\etc{\emph{etc}\onedot} 
 
\def\etal{\emph{et al}\onedot}
\makeatother
\newcommand{\PreserveBackslash}[1]{\let\temp=\\#1\let\\=\temp}
\newcolumntype{C}[1]{>{\PreserveBackslash\centering}p{#1}}
\newcolumntype{R}[1]{>{\PreserveBackslash\raggedleft}p{#1}}
\newcolumntype{L}[1]{>{\PreserveBackslash\raggedright}p{#1}}
\usepackage{xspace}
\makeatletter
\newcommand{\currentfsize}{\f@size pt}
\makeatother

\newdimen\fsize

\usepackage[capitalize]{cleveref}
\crefname{section}{Sec.}{Secs.}
\Crefname{section}{Section}{Sections}
\Crefname{table}{Table}{Tables}
\crefname{table}{Tab.}{Tabs.}

\makeatletter
\newcommand{\printfnsymbol}[1]{%
  \textsuperscript{\@fnsymbol{#1}}%
}
\makeatother
\usepackage[english]{babel}

\newcommand{\bc}{\mathbf{c}}
\newcommand{\bd}{\mathbf{d}}

\newcommand{\br}{\mathbf{r}}

\newcommand{\bx}{\mathbf{x}}

\newcommand{\nG}{\mathbb{G}}

\newcommand{\nP}{\mathbb{P}}

\newcommand{\nR}{\mathbb{R}}

\makeatletter
\def\@onedot{\ifx\@let@token.\else.\null\fi\xspace}
\def\eg{e.g\onedot} 
\def\ie{i.e\onedot} 
 
\def\etc{etc\onedot}

\def\etal{et~al\onedot}

\makeatother

\definecolor{glaucous}{rgb}{0.38, 0.51, 0.71}
\definecolor{stop_grad_red}{rgb}{0.75, 0, 0}
\definecolor{best}{rgb}{0.96, 0.57, 0.58}
\definecolor{second}{rgb}{0.98, 0.78, 0.57}
\definecolor{third}{rgb}{1.0, 1.0, 0.56}

\newcommand{\jx}[1]{\textcolor{black}{#1}}
\newcommand{\jxg}[1]{\textcolor{black}{#1}}

\begin{document}
\title{UC-NeRF: Uncertainty-aware Conditional Neural Radiance Fields from Endoscopic Sparse Views}
\author{Jiaxin Guo, Jiangliu Wang, Ruofeng Wei, Di Kang,\\ Qi Dou, \IEEEmembership{Member, IEEE}, and Yun-hui Liu, \IEEEmembership{Fellow, IEEE}
\thanks{This work is supported in part by Shenzhen Portion of Shenzhen-Hong Kong Science and Technology Innovation Cooperation Zone under HZQB-KCZYB-20200089, in part by the Research Grants Council of Hong Kong under Grant T42-409/18-R, Grant 14218322, and Grant 14207320, in part by the Hong Kong Centre for Logistics Robotics, in part by the Multi-Scale Medical Robotics Centre, InnoHK, and in part by the VC Fund 4930745 of the CUHK T Stone Robotics Institute. (Corresponding author: Yun-Hui Liu)}
\thanks{Jiaxin Guo, Jiangliu Wang are with CUHK T Stone Robotics Institute, The Chinese University of Hong Kong, Hong Kong, China. (Emails: jxguo@mae.cuhk.edu.hk, jlwang@cuhk.edu.hk)}
\thanks{Ruofeng Wei, Qi Dou are with the Department of Computer Science and Engineering, The Chinese University of Hong Kong, China. (Emails: ruofenwei2-c@my.cityu.edu.hk, qidou@cuhk.edu.hk)}
\thanks{Di Kang is with Tencent AI Lab, Shen Zhen, China. (Email: di.kang@outlook.com)}
\thanks{Yun-hui Liu is with CUHK T Stone Robotics Institute, The Chinese University of Hong Kong, and with Hong Kong Center for Logistics Robotics, Hong Kong, China (Email: yhliu@mae.cuhk.edu.hk)}
}
\maketitle

\begin{abstract}
Visualizing surgical scenes is crucial for revealing internal anatomical structures during minimally invasive procedures. Novel View Synthesis is a vital technique that offers geometry and appearance reconstruction, enhancing understanding, planning, and decision-making in surgical scenes. Despite the impressive achievements of Neural Radiance Field (NeRF), its direct application to surgical scenes produces unsatisfying results due to two challenges: endoscopic sparse views and significant photometric inconsistencies. In this paper, we propose uncertainty-aware conditional NeRF for novel view synthesis to tackle the severe shape-radiance ambiguity from sparse surgical views. The core of UC-NeRF is to incorporate the multi-view uncertainty estimation to condition the neural radiance field for modeling the severe photometric inconsistencies adaptively. Specifically, our UC-NeRF first builds a consistency learner in the form of multi-view stereo network, to establish the geometric correspondence from sparse views and generate uncertainty estimation and feature priors. In neural rendering, we design a base-adaptive NeRF network to exploit the uncertainty estimation for explicitly handling the photometric inconsistencies. Furthermore, an uncertainty-guided geometry distillation is employed to enhance geometry learning. Experiments on the SCARED and Hamlyn datasets demonstrate our superior performance in rendering appearance and geometry, consistently outperforming the current state-of-the-art approaches. Our code will be released at \url{https://github.com/wrld/UC-NeRF}.
\end{abstract}

\section{Introduction}

\begin{figure}[t]
\centering
\includegraphics[width=0.98\linewidth]{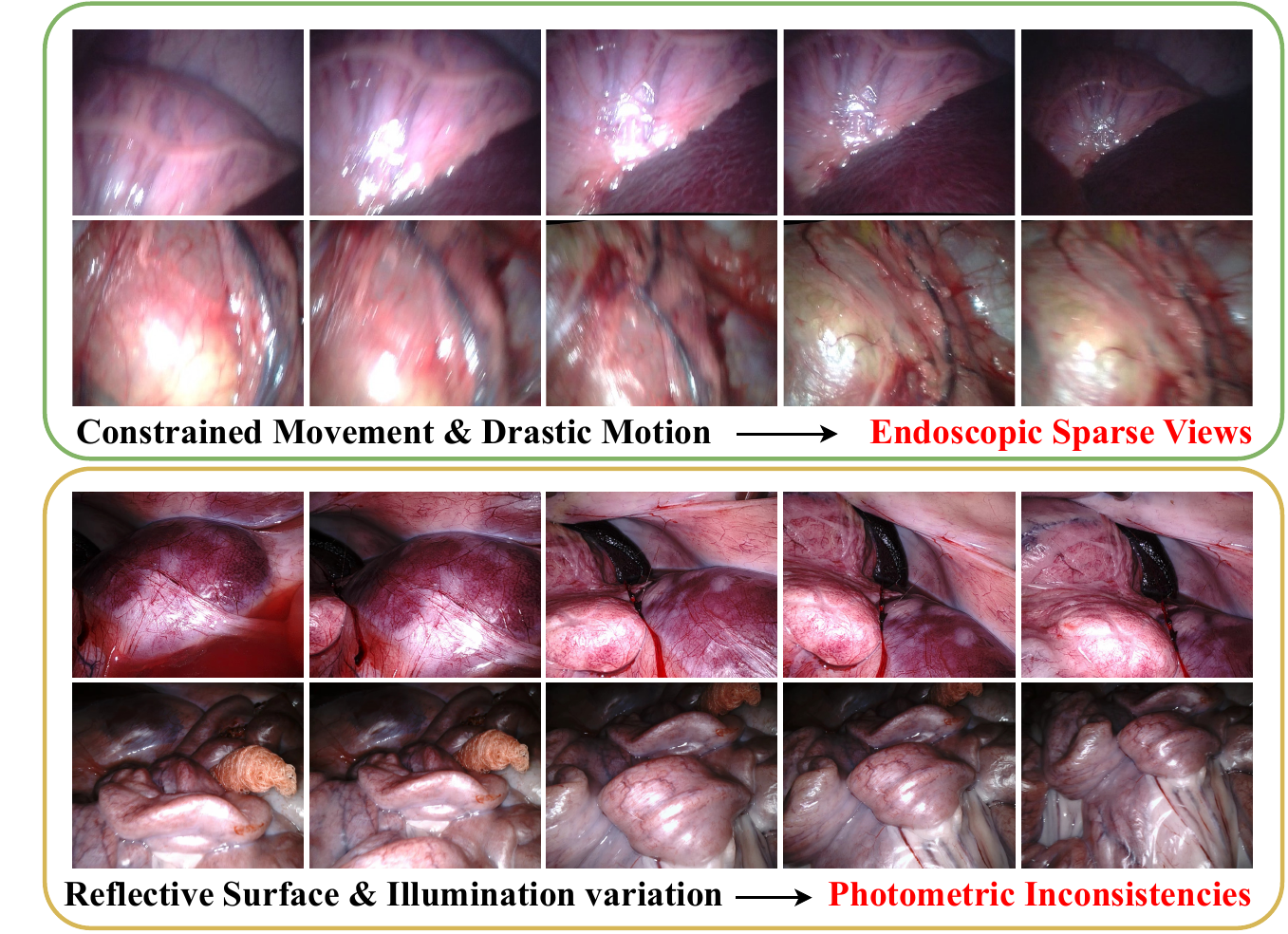}
\caption{\jx{\textbf{Employing NeRF~\cite{mildenhall2021nerf} in surgical scene} encounters two main challenges, \ie endoscopic sparse views and photometric inconsistency.}}
\label{fig:teaser_challenge}
\vspace{-15pt}
\end{figure}

Minimally invasive surgery (MIS) has achieved a significant advancement in modern surgical practices. 
It reduces surgical trauma, lessens post-operative discomfort, and shortens recovery time~\cite{bergen2014stitching}. 
The endoscope allows surgeons to inspect internal structures, facilitating precise navigation through complex anatomical landscapes~\cite{maier2013optical}. 
\jx{However, the 3D perception of the endoscope is impeded by the limited range of viewpoint changes inherent to endoscopic procedures, as well as the limited view area and two-dimensional imaging. }
\begin{figure*}[t]
\centering
\includegraphics[width=0.95\linewidth]{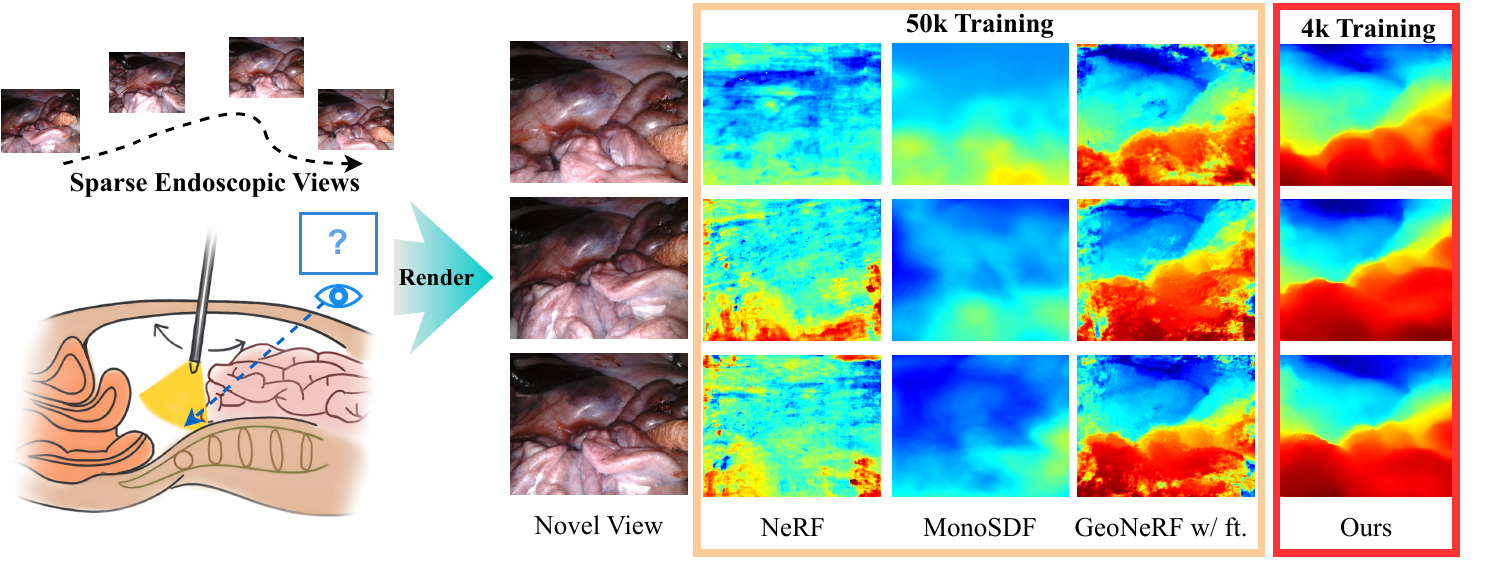}
\caption{
\textbf{Training NeRF on sparse surgical views is challenging.} 
NeRF~\cite{mildenhall2021nerf} fails to produce desirable views given sparse surgical scenes as inputs. 
State-of-the-art few-shot NeRF methods MonoSDF~\cite{yu2022monosdf} and GeoNeRF~\cite{johari2022geonerf} show degeneration in geometry rendering results. 
In contrast, our approach presents consistent improvement and achieves faster convergence in 4k compared to the 50k optimization of other baselines.}
\label{fig:teaser}
\vspace{-8pt}
\end{figure*}
These limitations make it challenging for surgeons to perceive depth and fully assess the surrounding conditions, which can hinder their understanding of the internal anatomy and potentially impact the effectiveness of surgical interventions. 
Additionally, traditional visualization systems in surgery, such as Ultrasound, Magnetic Resonance Imaging (MRI), or Computed Tomography (CT), increase the cost and complexity of medical imaging~\cite{malhotra2023augmented}. 
While 3D reconstruction methods allow surface geometry learning, they fall short of predictive ability and flexibility in exploring and visualizing surgical scenes from different perspectives with high-fidelity details.

Utilizing endoscopic multi-view images as input, the novel view synthesis technique generates photo-realistic free-view images of surgical scenes and intricate abdominal structures. 
This technique provides advantages for various applications, including virtual reality interactions, intra-operative surgical navigation, and autonomous robotic surgery~\cite{tang2018augmented, penza2017envisors, chen2018slam}. 
In this area, the Neural Radiance Fields (NeRF) approach~\cite{mildenhall2021nerf} has shown remarkable success. 
It synthesizes novel viewpoints from dense sets of input images using implicit volumetric representations, paving the way for an era of visually immersive and interactive surgical practices.

While NeRF has shown remarkable effectiveness in handling natural images, its application in surgical scenes often results in subpar rendering, due to two challenges: endoscopic sparse views and photometric inconsistency, as illustrated in Fig. \ref{fig:teaser_challenge}. \jx{Endoscopic sparse views are primarily due to constrained camera movement and drastic motion. The movement of the endoscope is inherently restricted by the narrow internal spaces within the body and the limited flexibility of the endoscope. These constraints confine the camera to a limited number of views, making it difficult to observe the same spot from different viewpoints. Moreover, when surgeons maneuver the endoscope, drastic movements under low camera frame rates can cause intermittent visibility and motion blur, hindering the acquisition of clear and consecutive frames. Both two factors reduce the overlap between captured views, leading to the sparsity of useful visual data.} This is particularly challenging for NeRF, which relies on dense views for accurate 3D geometry inference and novel view synthesis. Besides, surgical scenes frequently exhibit considerable photometric inconsistencies, due to non-Lambertian reflections and fluctuating illumination. Such inconsistencies complicate the accurate capture of scene geometry and appearance, leading to potential artifacts and incorrect density distributions in NeRF synthesized views due to its fundamental design around minimizing RGB error. These factors contribute to a pressing question: \textit{How can we enhance NeRF's ability to handle the severe shape-radiance ambiguity problem caused by the endoscopic sparse views and photometric inconsistency?}

 \jx{To address this problem, we take the inspiration from few-shot NeRF to solve the problems induced by endoscopic sparse view.
Some methods are proposed for few-shot NeRF to enhance the rendering performance and reduce the shape-radiance ambiguity, which could be roughly categorized into two classes: NeRF with pre-training~\cite{yu2021pixelnerf, chen2021mvsnerf, johari2022geonerf} and NeRF with geometry guidance~\cite{roessle2022dense, wei2021nerfingmvs, niemeyer2022regnerf, yu2022monosdf, deng2022depth}. }
 The former class requires large datasets to pre-train the generalizable NeRF and fine-tune it on similar target scenes. 
 The latter class employs geometric information (depths, normals, pointclouds) to supervise and regularize the neural rendered depth or control the ray sampling range to get rid of outliers. While these methods achieve improvements compared to NeRF, they ignore the extent of photometric inconsistencies and directly inject the constraints equally in spatial, leading to unsatisfying performance as presented in Fig.~\ref{fig:teaser}.
 To tackle the severe shape-radiance ambiguity in surgical scene, we aim to explicitly detect the photometric inconsistency from the multi-view inputs and enable the neural radiance fields to adaptively model the regions with the uncertainty estimation.

In this paper, we propose a new network for surgical novel view synthesis, \jx{to empower NeRF with great robustness and efficiency to tackle the challenging surgical scene with inconsistencies and sparse views, i.e. UC-NeRF. }
The key novelty of our UC-NeRF lies in incorporating the multi-view uncertainty information with geometry and appearance priors to condition the neural radiance field \jx{for improved accuracy and robustness to sparse views}.
Our network has three essential designs to exploit the uncertainty information: 
i) A \textbf{consistency learner} to build the geometry correspondence and learn the uncertainty estimation across sparse multi-views. 
ii) An \textbf{uncertainty-aware dual-branch NeRF} designed with base-adaptive architecture utilizing the uncertainty information to handle photometric inconsistencies and solve shape-radiance ambiguity. 
iii) The \textbf{distillation from monocular geometry priors} to optimize the neural rendering with the uncertainty guidance to improve the accuracy in rendered depth. We validate our proposed approach on the SCARED~\cite{allan2021stereo} and Hamlyn datasets~\cite{mountney2010three, stoyanov2010real, pratt2010dynamic}. The extensive experiments demonstrate the state-of-the-art performance of UC-NeRF in novel view synthesis from sparse endoscopic views, with consistent improvement in effectiveness and efficiency.

In summary, our contributions are three folds: 1) To the best of our knowledge, we are the first to present a NeRF-based method addressing the challenging problem of novel view synthesis from endoscopic sparse images. 2) We devise an uncertainty-aware dual-branch NeRF to exploit the learned uncertainty information, to recover view-dependent appearance while reducing the shape radiance ambiguity, with the distillation from the monocular geometry priors to enhance the accuracy and robustness. 
\jx{3) The experiment results demonstrate that our method outperforms the previous state-of-the-art baselines, showing the superior efficiency and robustness to endoscopic sparse views.}

\section{Related Works}

\noindent\textbf{Novel View Synthesis.}
Novel view synthesis focuses on generating new images or views of a scene from viewpoints not captured in the original imagery. 
Traditionally, novel view synthesis is based on geometric methods like image-based rendering~\cite{mcmillanunstructured, sinha2009piecewise, chaurasia2013depth} and light field rendering~\cite{levoy1996light, wood2000surface, chen2018deep, srinivasan2017learning}, which require precise camera calibration and struggle with complex scenes. The integration of deep learning allows a significant advancement for more realistic view generation~\cite{choi2019extreme, mildenhall2019local, srinivasan2019pushing}. A major breakthrough in this area was the development of NeRF~\cite{mildenhall2021nerf}, which used a fully connected deep network to model volumetric scene functions, greatly enhancing the quality of synthesized views. Following NeRF, various extensions have emerged by improving speed, quality, and generalization~\cite{chen2021mvsnerf, yu2021pixelnerf, yu2022monosdf, niemeyer2022regnerf, wei2021nerfingmvs}.
\begin{figure*}[t]
\centering
\includegraphics[width=1.0\linewidth]{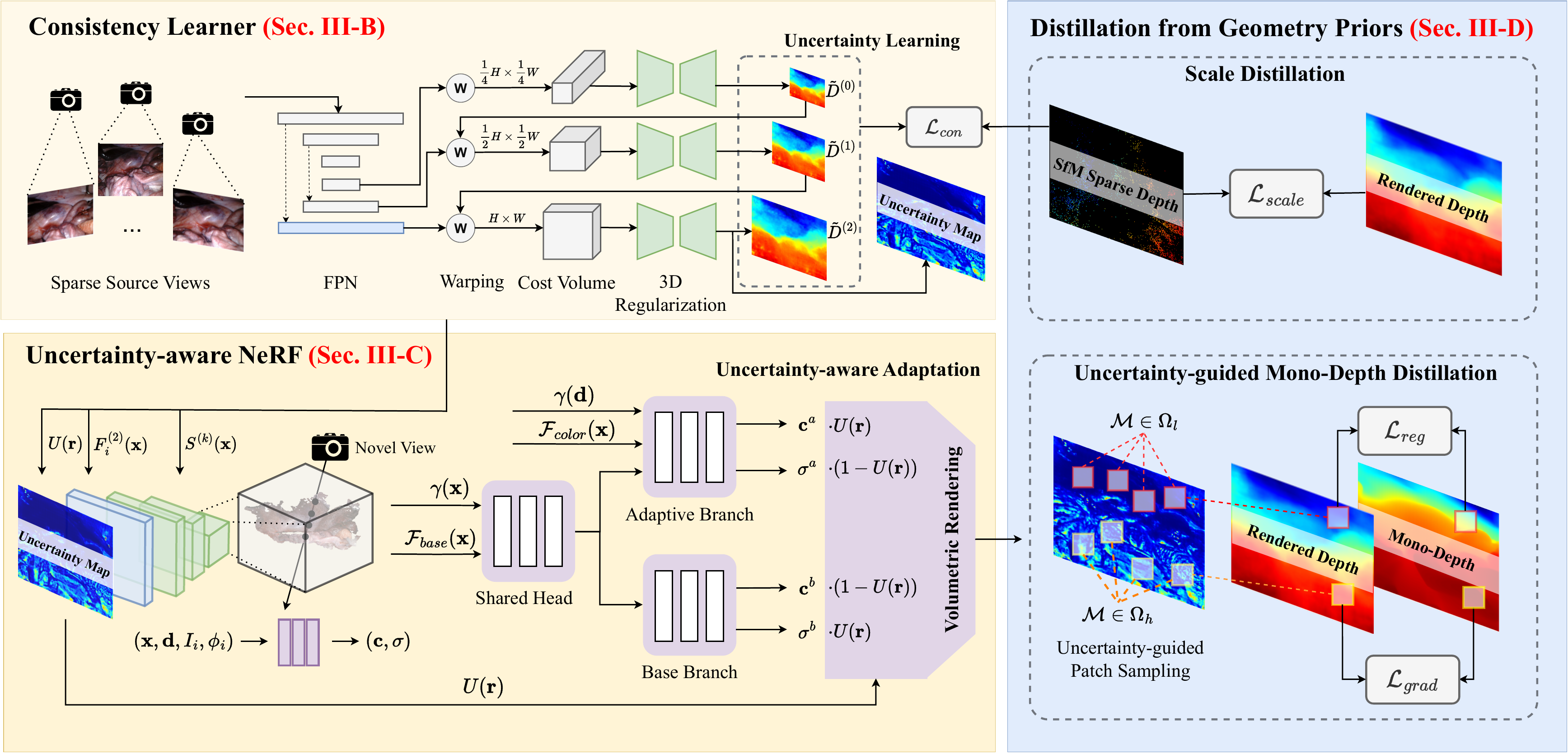}
\vspace{-8pt}
\caption{\textbf{Overview of our Uncertainty-aware Conditional NeRF (UC-NeRF).} 
We first build a consistency learner upon the multi-view stereo network, to capture the view-consistent constraints to generate the uncertainty map. 
Then, the uncertainty-aware NeRF takes image features (from FPN and 3D regularization module) and the uncertainty map as input to predict the radiance field, resulting in reduced shape-radiance ambiguity and improved rendering accuracy.
Finally, we introduce the distillation from geometry priors for further optimizing the neural rendering results.
}
\label{fig:network}
\vspace{-10pt}
\end{figure*}
\\
\noindent\textbf{NeRF from Surgical Scenes.}
While NeRF has shown its potential in novel view synthesis, current approaches focus on 3D reconstruction using neural implicit fields from stereo endoscope videos. 
EndoNeRF~\cite{wang2022neural} explores neural rendering for deformable tissue reconstruction from stereo endoscope inputs and devised a mask-guided ray-casting strategy to address the tool occlusion challenge. Sun \etal~\cite{sun2023dynamic} propose a depth estimation network and a reconstruction network utilizing neural radiance fields for dynamic reconstruction. EndoSurf~\cite{zha2023endosurf} proposes to model the deformation, geometry, and appearance separately to represent the deforming surfaces. Unlike these approaches that focus on single-view stereo video reconstruction, to the best of our knowledge, we are the first to present a sparse-view NeRF approach to tackle the challenge of novel view synthesis from endoscopic sparse views.

\noindent\textbf{Few-shot Neural Rendering.}
Recently, the few-shot NeRF techniques that utilize sparse views for novel view synthesis have opened up new possibilities for improving the robustness and efficiency of NeRF, which can be classified into two classes: NeRF with pre-training and NeRF with geometry guidance.
NeRF with pre-training in the first category pretrain a generalizable NeRF across multiple large datasets before fine-tuning on specific targets. PixelNeRF~\cite{yu2021pixelnerf} leverages CNN features from input images to predict a continuous neural scene representation. IBRNet~\cite{wang2021ibrnet} introduces a network architecture to estimate radiance and volume density with appearance information from multiple source views. MVSNeRF~\cite{chen2021mvsnerf} takes usage of cost volumes to reason the prior features to enhance the neural radiance field reconstruction. GeoNeRF~\cite{johari2022geonerf} incorporates a transformer-based attention mechanism with volume rendering to manage complex occlusion conditions. However, these methods are prone to degeneration during significant data domain shifts, such as surgical scenes that differ vastly from general ones. Moreover, the need for extensive and diverse pre-training datasets results in substantial time and computational costs. In this paper, we advocate the conditional NeRF by exploiting features and uncertainty estimation from sparse source images and realize generalizable training on the multiple surgical scenes, outperforming previous methods even without fine-tuning.
NeRF with geometry guidance focuses on guiding NeRF training with geometric information. For example, NerfingMVS~\cite{wei2021nerfingmvs} takes MVS depth from COLMAP~\cite{schonberger2016structure} as a prior and employs a monocular depth network to guide NeRF optimization. RegNeRF~\cite{niemeyer2022regnerf} regularizes geometry and appearance from unobserved viewpoints, refining ray sampling space over time. DS-NeRF~\cite{deng2022depth} uses sparse 3D points from Structure-from-Motion (SfM) to monitor and adjust NeRF ray termination, enabling faster training with sparse views. MonoSDF~\cite{yu2022monosdf} integrates multiple monocular geometric priors into neural implicit surface reconstruction. Roessle \etal~\cite{roessle2022dense} upsample SfM sparse points into dense depth maps to guide NeRF optimization. SparseNeRF~\cite{wang2022sparsenerf} employs a local depth ranking regularization and a spatial continuity regularization to distill the depth priors. ConsistentNeRF~\cite{hu2023consistentnerf} learns the 3D consistency by leveraging depth information to regularize the multi-view and single-view among pixels. Despite these advances, they ignore the extent of photometric inconsistencies and inject the constraints equally in spatial. Therefore, the injected hard geometry constraints easily cause over-fitting in sparse training data, leading to degeneration in test novel views. Our proposed UC-NeRF introduces an uncertainty-aware conditional NeRF to address these challenges, by leveraging the uncertainty information to model the 3D surgical scene and guide the optimization.

\noindent\jx{\textbf{Uncertainty Estimation in NeRF.} Uncertainty estimation has been adopted in diverse areas of computer vision to improve the interpretability and reduce the risk of the model. NeRF-W~\cite{martin2021nerf} leverages the uncertainty to tackle the transient object problem. S-NeRF~\cite{shen2021stochastic} learns a probability distribution to quantify the uncertainty associated with the scene information. CF-NeRF~\cite{shen2022conditional} introduces latent variable modeling and conditional normalized flow to incorporate uncertainty quantification into NeRF. ActiveNeRF~\cite{pan2022activenerf} incorporates the uncertainty estimation into a NeRF model by modeling the radiance values as a Gaussian distribution. In this paper, we aim to address the challenge of novel view synthesis from endoscopic sparse views. Unlike previous works, we incorporate the multi-view uncertainty from the consistency learner with dual branch NeRF, to explicitly handle the photometric inconsistencies from endoscopic sparse views.}

\section{Uncertainty-aware Conditional NeRF}

Given a set of sparse input images $\{ I_1, \dots, I_N\}$ and their corresponding camera parameters $\{ \phi_1, \dots, \phi_N\}$ as input,
our goal is to reconstruct a radiance field that can faithfully capture the view-independent effects and the underlying true geometry,
so that we can volume-render an image from any novel viewpoint and thus facilitate the diagnosis.
Mathematically, this process is denoted as follows:
\begin{equation}
    (\bc, \sigma)  = \text{UC-NeRF}(\gamma(\bx), \gamma(\bd); I_i, \phi_i),
\end{equation}
where $\gamma$ is the encoding function to map position $\bx$ and view direction $\bd$ to a higher dimensional space.
Through conditional inputs, our method enables 
generalizable training across multiple surgical scenes, thereby promoting training efficiency and robustness. 

As in Fig.~\ref{fig:network}, UC-NeRF contains two major components: 1) a consistency learner that builds the geometry correspondences across multi-view inputs and generates an uncertainty map (Sec.~\ref{sec: consist_learner}); and 2) a conditional NeRF that enables uncertainty-aware neural radiance fields reconstruction (Sec.~\ref{sec: c_nerf}).
To maximize generalization capability and reduce errors caused by shape-radiance ambiguity, 
we propose to distill geometric priors from the estimated sparse SfM points and a monocular depth estimator into our UC-NeRF (Sec.~\ref{sec: distillation}), resulting in further improved robustness and accuracy in the rendered depth.

\subsection{Preliminaries}

We first briefly introduce some NeRF basics~\cite{mildenhall2021nerf} which are used in this paper.
NeRF takes as input a set of posed images and represents a scene as a continuous volumetric function parameterized by MLPs. 
Given a 3D point $\bx \in \nR^3$ and a viewing direction $\bd \in \nR^2$, NeRF learns to map from $(\bx, \bd)$ to the volume density $\sigma$ and the emitted color $\bc=(r,g,b)$:
\begin{equation}
    (\bc, \sigma) = \text{MLP}(\gamma(\bx), \gamma(\bd)).
\end{equation}
\jx{The color of an image pixel is calculated with the volume rendering \cite{mildenhall2021nerf}.
Specifically, a ray $\bold{r}(t) = \bold{o} + t \bold{d}$ is determined by the camera origin $\bold{o}$ and the pixel location $\bold{p}$, where $\bold{d}$ is the unit direction vector passing from the camera origin $\bold{o}$ to the pixel $\bold{p}$, $t$ is the distance of a sampling point to the origin on this ray.
The volume rendering equation~\cite{mildenhall2021nerf} to obtain a pixel's color is defined as follows:
\begin{equation}\label{eq:C_render}
    \begin{aligned}
        \hat{C}(\bold{r}) &= \sum_{i=1}^N T_i(1-\text{exp}(-\sigma_i\delta_i))\bold{c}_i, \\
    \text{where }T_i &= \text{exp}(-\sum_{j=1}^{i-1}\sigma_j\delta_j), \delta_i = t_{i+1} - t_i,
    \end{aligned}
\end{equation}
where 
$N$ is the number of sample points along each ray, 
$\sigma_i$ is the density value of point $\bold{x}_i$,
$\delta_i$ is the distance between two consecutive sample points along the ray, 
$T_i$ is the accumulated transparency from the camera origin. Following~\cite{mildenhall2021nerf, roessle2022dense}, the depth of an image pixel can be similarly calculated by integrating every sample point's distance to the camera origin:
\begin{equation}\label{eq:D_render}
\hat{D}(\bold{r}) = \sum_{i=1}^N T_i(1-\text{exp}(-\sigma_i\delta_i))t_i. 
\end{equation}}
With the above differential volume rendering process, NeRF optimizes the radiance fields by minimizing the reconstruction error between the rendered color and the ground truth color:
\begin{equation}
\mathcal{L}_{rgb} = \parallel \hat{C}(\br) - C(\br) \parallel^2_2.
\end{equation}
However, NeRF reconstruction degrades drastically, especially the geometry part, due to the sparse endoscopic views and the photometric inconsistencies across images caused by varying lighting conditions during the surgical operation.

\subsection{Consistency Learner}\label{sec: consist_learner}

In UC-NeRF, a consistency learner is utilized to exploit robust geometric information from sparse SfM depth and learn the consistency.
Specifically, it adopts the CasMVSNet~\cite{gu2020cascade} 
as the backbone to extract intermediate image features and construct cascade cost volumes to predict a dense depth map and an uncertainty map.
The 2D image features and the 3D neural volumes, which contain geometry and appearance information of the target surgical scenes, are used as the condition input of the later NeRF network (detailed in \jx{Sec.}~\ref{subsec: ba_render}).
The uncertainty map is used to adaptively re-weight the predictions before radiance integral (detailed in \jx{Sec.}~\ref{Subsec: uncertain_adapt}).
With the help of this consistency learner, our UC-NeRF can use better conditional information as input and estimate more accurate novel view images and depth maps.
\subsubsection{Cascade Neural Volumes}
Given $N$ sparse source views $\{ I_{i}\}_{i=1}^N$ with resolution size $H \times W$ as input, we utilize a Feature Pyramid Network (FPN)~\cite{lin2017feature} to extract image features in different spatial resolutions across three stages.
\begin{equation}
    F_{i}^{(k)} = \text{FPN}(I_i), \quad  k=\{0, \, 1, \, 2\},
\end{equation}
where $F_{i}^{(k)} \in \nR^{\frac{H}{2^{2-k}} \times \frac{W}{2^{2-k}}\times 2^{-k}Z}$ represents the 2D feature maps extracted at the stage $k$ from the $i_{th}$ input view, $Z$ is the feature dimension of stage $0$.

Next, we warp the 2D feature maps from different source views to the plane sweeping volume feature $ V_{i}^{(k)}$ on the frustum of the target view following~\cite{yao2018mvsnet, gu2020cascade}. 
Given the camera parameters $\{ \boldsymbol{\phi}_i \}_{i=1}^N$ for the input source images, and $\boldsymbol{\phi}_0$ for the target view, we can apply homography warping to warp the 2D feature maps $F_{i}^{(k)}$ from source views into hypothetical planes of the target view, forming 3D features $ V_{i}^{(k)}$. Note that the target view is the novel view to be rendered, which is the same as the reference camera in MVS methods~\cite{yao2018mvsnet}.
Following CasMVSNet~\cite{gu2020cascade}, the hypothesis depth planes range from $d_{max}^{(k)}$ to $d_{min}^{(k)}$ with $Y^{(k)}$ discrete depth values evenly spaced in-between.
The depth planes are configured in a coarse-to-fine manner from stage $0$ which uses the largest depth range and the most planes, to stage $2$.

The cost volume is calculated with a variance-based metric from 3D feature $V_{i}^{(k)}$~\cite{yao2018mvsnet, gu2020cascade}.
Note that we only use features from the neighboring source views during feature extraction and the following depth estimation, since we do not have access to the image of the novel view to be synthesized. 
We further regularize the cost volume with a 3D-CNN to generate a 3D neural volume $S^{(k)}$ and a probability volume $P^{(k)}$:
\begin{equation}
\begin{aligned}
S^{(k)}, P^{(k)} = \text{3D-CNN}(\text{Var}(V_{i}^{(k)})),
\end{aligned}
\label{eq:depth_prob}
\end{equation}
where $S^{(k)} \in \nR^{Y^{(k)} \times \frac{H}{2^{2-k}} \times \frac{W}{2^{2-k}} \times 2^{-k} Z}$ denotes the 3D neural volume storing the geometry information of the target view,
and is used as one conditional input to build the neural radiance fields. 
\jxg{$P^{(k)}\in \nR^{Y^{(k)} \times \frac{H}{2^{2-k}} \times \frac{W}{2^{2-k}}}$ represents the probability volume specifying every spatial location's depth probability among all possible depth planes. After computing $P^{(k)}$, we also perform a softmax operation on the depth plane dimension of $P^{(k)}$ to ensure that the probability of the depth hypothesis remains in the range $[0, 1]$.}
The final depth estimation $\Tilde{D}^{(k)}$ of the target view is obtained as the expectation of different depth hypothesis~\cite{yao2018mvsnet}:
\begin{equation}\label{eq: mvs_depth}
\Tilde{D}^{(k)} = \sum_{d=d_{min}^{(k)}}^{d_{max}^{(k)}}d \times P^{(k)}(d).
\end{equation}
We further enhance the consistency learner with view-consistent points from SfM. 
Specifically, we utilize the sparse depth $D_{sfm}$ projected from SfM points in every view to supervise the predicted depth from the consistency learner. 
Taking into account existing noise, we use the SfM reprojection error $\omega$ as the weight in the sparse depth for the regression:
\begin{equation} \label{eq: con}
\begin{aligned}
    L_{con} &= \sum_{k=0}^{k=2} \alpha^{(k)}  \text{exp}(-(\omega / \Bar{\omega})^2) \parallel  \Tilde{D}^{(k)} - D_{sfm} \parallel_1 , \\
\end{aligned}
\end{equation}
where $ \text{exp}(-( \omega / \Bar{\omega})^2)$ is the weight for depth loss at stage $k$, which injects larger weight on the depth loss for the point with a smaller reprojection error. Given that SfM leverages SIFT points and bundle adjustment optimization, it excels in accurately capturing geometric correspondences. Consequently, the consistency learner can exploit the reliable geometric correspondences captured by SfM, \ie light-invariant and visible points among the source view inputs.
\begin{figure}[t]
\centering
\includegraphics[width=0.98\linewidth]{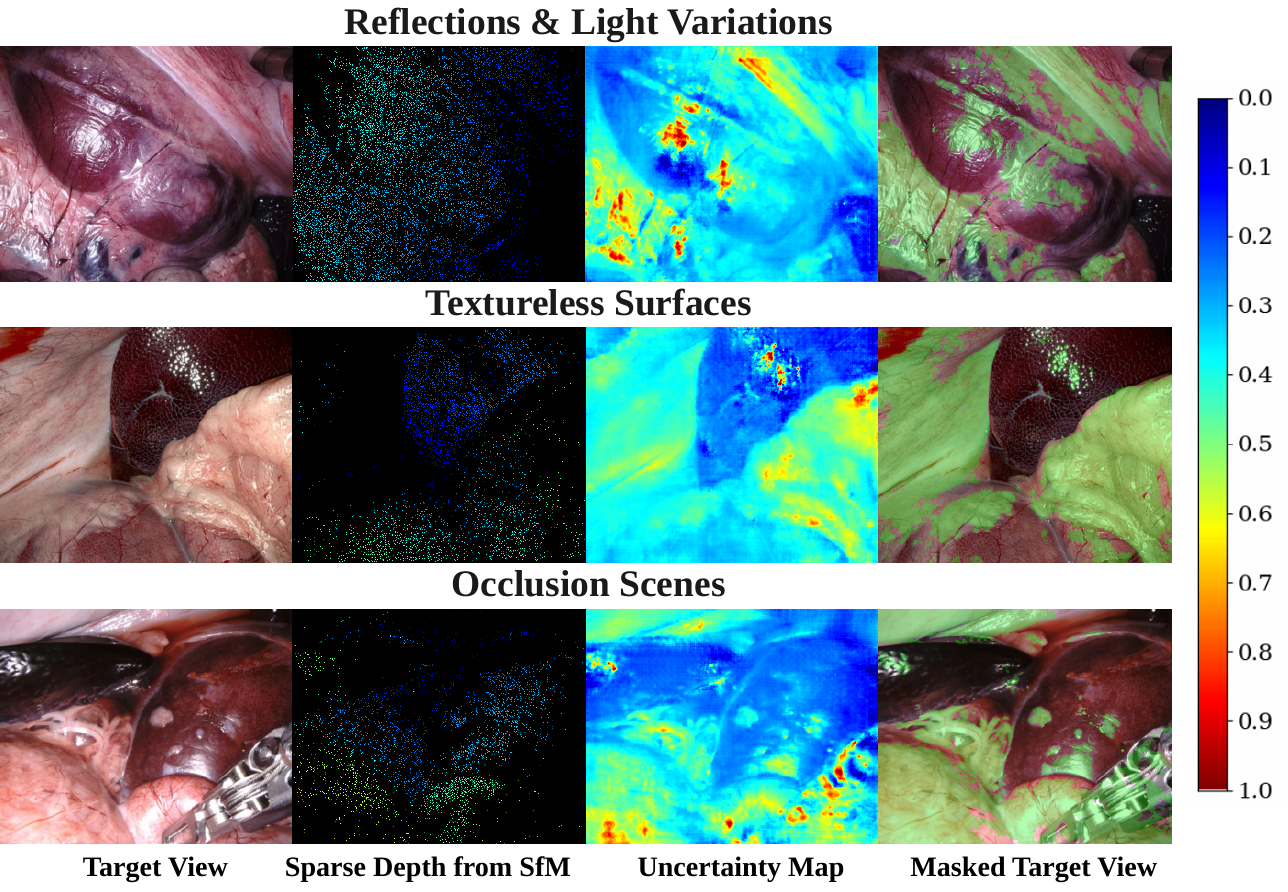}
\caption{\jxg{\textbf{Visualization of the Sparse SfM depth and the estimated uncertainty map.} With the guidance from SfM, the uncertainty map measures the extent of photometric inconsistency. The masked target view indicates the region with uncertainty larger than the mean value.}}
\label{fig: uncertainty}
\vspace{-10pt}
\end{figure}

\subsubsection{Uncertainty Learning}
To model the extent of the photometric inconsistencies, an uncertainty map of the target view is generated according to the probability distribution along the depth hypothesis. \jx{Specifically, for every pixel ($u, v$), its uncertainty is calculated according to four probability values from the probabilistic volume $P^{(2)}$ that are closest to its depth estimation ($\Tilde{D}^{(2)}$ in Eq. (8)) following~\cite{yao2018mvsnet, gu2020cascade}:}
\begin{equation}
    \begin{aligned}
        U(u, v) = 1 - \sum_{i=j-1}^{j+2} P^{(2)}(u, v, i) \bigg|_{j=\text{Index}(\Tilde{D}^{(2)}(u, v))},
    \end{aligned}
\end{equation}
where $\text{Index}(\cdot)$ indicates the index of the hypothesis depth plane. If a point exhibits consistency across multiple views, it is prone to have an unimodal distribution in its depth estimation probability. This results in the estimated probability $P^{(2)}$, being close to the peak of this unimodal distribution, indicating low uncertainty. 
In contrast, 
multimodal distribution signifies higher uncertainty, as there is a lack of consensus among the different views on the correct depth value. As the uncertainty map shown in Fig.~\ref{fig: uncertainty}, regions with high uncertainty tend to have minimal texture, large occlusion, and severe reflective surfaces. This suggests that our consistency learner is able to produce a sensible measure of uncertainty based on the guidance of SfM sparse points.

\subsection{Uncertainty-aware dual-branch NeRF}\label{sec: c_nerf}
We propose a dual-branch NeRF utilizing the uncertainty information to explicitly handle photometric inconsistencies caused by moving light and non-Lambertian surfaces.
Specifically, our dual-branch NeRF contains a base branch $\text{MLP}_{\theta_b}$ spatially weighted by the confidence score (i.e. 1 - uncertainty score), which aims at modeling view-consistent appearance and geometry in the surgical scene, and an adaptive branch $\text{MLP}_{\theta_a}$ weighted by the uncertainty score, which aims at modeling view-dependent effects and details, \ie illumination variations, non-Lambertian texture.
The reason is that regions with higher uncertainty values from the depth estimator usually lack consistency across neighboring multi-views, which are supposed to be modeled by the adaptive branch.

\subsubsection{Base-Adaptive Rendering Network}\label{subsec: ba_render}
The NeRF is conditioned on the feature priors from the consistency learner to generalize in different scenes with better efficiency and rendering performance (see Fig.~\ref{fig: efficiency}).
Concretely, we directly build the cost volume upon the target (i.e. novel) view frustum to collect more ``original'' appearance and geometry features from the target view rather than highly processed abstract features from other views. This is different from MVSNeRF~\cite{chen2021mvsnerf}, which builds cost volume upon the neighboring input view and then warps the 3D points to this different view for sampling condition features.
This modification of the cost volume reconstruction process results in better geometry accuracy and improved image rendering performance (See Tab.~\ref{table: scared_comparison}).

Specifically, a sample point is warped to input views to obtain appearance prior (i.e. $\{I_i(\bx_{0\rightarrow i})\}_{i=1}^N$)
and geometry prior $S^{(k)} \{(\bx)\}_{k=0}^2$
.
Note that $S^{(k)}$ is a 3D volume space in NDC coordinate frame and we use trilinear interpolation. Mathematically, the condition input $\mathcal{F}_{base}(\bx)$ and the conditional NeRF are defined as: 
\begin{align}
\mathcal{F}_{base}(\bx) &= \text{Concat}(\{ S^{(k)} (\bx)\}_{k=0}^2, \{I_i(\bx_{0\rightarrow i})\}_{i=1}^N),\\
 h &= \text{MLP}_{\theta_s}(\gamma(\bx), \mathcal{F}_{base}(\bx)),
\end{align}
where $h$ is the shared latent feature used by both base and adaptive branches.
\begin{align}
\bc^{b}, \sigma^{b} &= \text{MLP}_{\theta_{b}}(h),\\
\bc^{a}, \sigma^{a} &=  \text{MLP}_{\theta_{a}}(h, \gamma(\bd), \mathcal{F}_{color}(\bx)),  \\
\text{where} \quad \mathcal{F}_{color}(\bx) &= \{ F_{i}^{(2)}(\bx_{0\rightarrow i}) \}_{i=1}^N.
\end{align}
The density $\sigma^{b}$ and color $\bc^{b}$ of the base branch is decoded solely by $h$ without viewing vector input since this branch is designed to model the underlying true geometry and diffuse colors.
As for the adaptive branch, both its density $\sigma^{a}$ and color $\bc^{a}$ are dependent on view direction $\gamma(\bd)$ since it is designed to model the inconsistencies.
Note that we also use image feature as input since we find it is helpful experimentally possibly due to its robustness, \ie deep features with large enough context information. 
Specifically, we use the features $\{ F_{i}^{(2)}(\bx_{0\rightarrow i}) \}_{i=1}^N$ from the last layer of the FPN network.

\subsubsection{Uncertainty-aware Adaptation} \label{Subsec: uncertain_adapt}
Since the uncertainty measure reflects the geometric reliability and photometric inconsistency of each point,
it is suitable to be used as weight to balance and control the contribution of the two branches.
Specifically,
radiance fields predicted by the two branches are spatially weighted summed according to uncertainty score $U(\br)$ as follows:
\begin{align}
    \bc &= \bc^{b} \cdot (1-U(\br)) + \bc^{a} \cdot U(\br), \\
     \sigma &= \sigma^{b} \cdot U(\br) + \sigma^{a} \cdot (1-U(\br)).
\end{align}
The final image is volumetric rendered following Eq. (\ref{eq:C_render}) and Eq. (\ref{eq:D_render}).
\jxg{Under this setting, in color prediction, the adaptive branch $\bc^{a}$ contributes more to regions with higher uncertainty, capturing view-dependent photometric effects that vary significantly with different viewing angles. The base branch $\bc^{b}$ is more influential in regions with higher confidence, where the appearance is stable and predictable. 
Conversely, in density prediction, the base branch $\sigma^{b}$ is dominant in regions with higher uncertainty. This helps maintain geometric reliability and multi-view consistency, avoiding shape-radiance ambiguity which can lead to degenerated reconstructions. For regions with higher confidence and fewer directional variations, the adaptive branch $\sigma^{a}$ could contribute more, providing additional textures and details to enhance the reconstruction quality (See Tab.~\ref{table: ablation}). } 

\noindent\textbf{Discussion.} 
Following this adaptation, the base branch provides reliable and accurate geometry information to reduce the uncertainty and improve the robustness for these rays (See Fig.~\ref{fig: abl_compo}). In color rendering, sampled rays with higher uncertainty, prone to complex photometric effects, are weighted towards the adaptive branch to compensate for more view-dependent details (See Fig.~\ref{fig: branch_vis}).  
With the uncertainty map modulating the base and adaptive branches, our UC-NeRF is enabled to construct the neural radiance field with spatial uncertainty, facilitating more stable and controllable learning to solve the shape-radiance ambiguity. With the balance modulated in the two branches, the differential uncertainty-aware adaptation is capable of synchronizing the uncertainty learned by multi-view stereo with the neural rendering process.

\subsection{Distillation from Monocular Geometry Priors}\label{sec: distillation}

To further improve the geometry consistency, we exploit the geometry priors from monocular images to guide the training of UC-NeRF for scale-aware depth learning. We employ a two-fold approach: 1) Scale distillation from SfM. 2) Uncertainty-guided monocular depth distillation.

\begin{figure*}[t]
\centering
\includegraphics[width=0.9\linewidth]{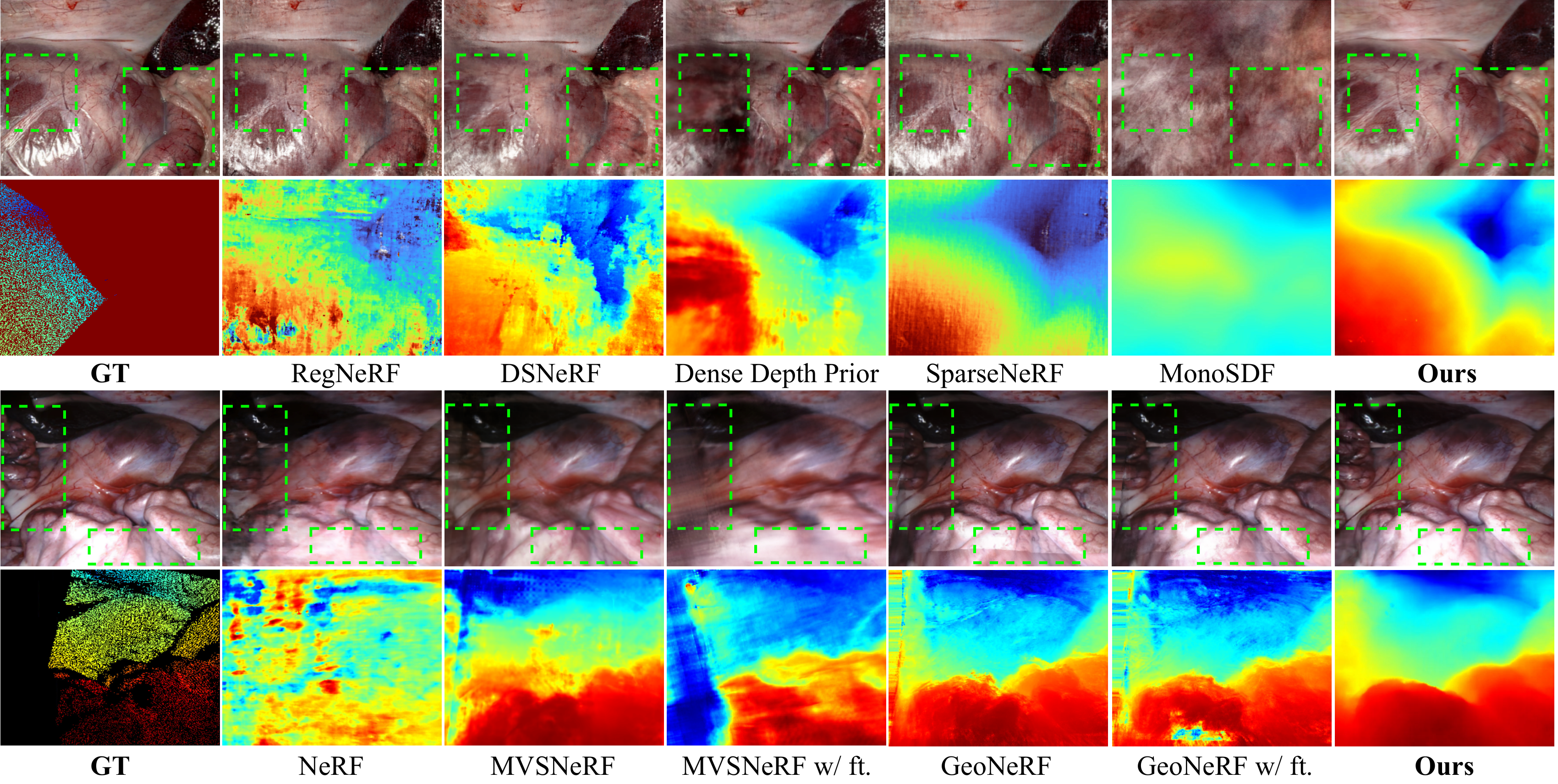}
\caption{\textbf{Qualitative Comparisons of rendered color and depth}. Given sparse input views, existing approaches show rendering results with blur and artifacts, suffering from photometric inconsistency. Our UC-NeRF can generate fine-grained details and consistent depth.}
\label{fig: qualitative}
\vspace{-5pt}
\end{figure*}

\subsubsection{Scale Distillation}
To preserve the scale consistency among sparse views, We first incorporate the sparse depth from SfM, which intrinsically preserves real-world scale. This sparse depth is used to guide our method in learning the correct scale of the scene through a scale distillation Loss. \jx{Similar to  Eq. (\ref{eq: con}), to alleviate the negative influence of inaccurate depth, we adopt the weight $\text{exp}(-( \omega / \Bar{\omega})^2)$ to suppress the supervision from unreliable depth estimations (i.e. with larger reprojection errors):}
\begin{equation}
\begin{aligned}
\mathcal{L}_{scale} &= \sum_{\br \in \mathcal{R}} \text{exp}(-(\omega / \Bar{\omega})^2) \parallel \hat D(\br) - D_{sfm}(\br) \parallel_1,
\end{aligned}
\end{equation}
where $\mathcal{R}$ denotes the sampled set of pixels in the region where SfM depth is available. We minimize the $L_1$ loss between the rendered depth $\hat D(\br)$ and the SfM sparse depth $D_{sfm}$, with SfM reprojection error $\omega$ to weight the loss.

\subsubsection{Uncertainty-guided Mono-Depth Distillation}To optimize the region where SfM is sparse or unavailable, we leverage a monocular depth estimation model, Dense Prediction Transformer (DPT)~\cite{ranftl2021vision}, to guide our UC-NeRF. We introduce an uncertainty-guided mono-depth distillation, taking the uncertainty map as a reference to employ different losses spatially for depth supervision to sampled patches.

We first sample image patches in the region with high uncertainty denoted as $\Omega_h$. Since the constraint of SfM sparse point is not enough for the high uncertainty region $\Omega_h$, we apply a scale-invariant depth gradient loss $\mathcal{L}_{grad}$ to supervise the gradient difference between rendered depth $\hat D$ with the monocular depth $D_{dpt}$. 
\begin{equation}
\begin{aligned}
    \mathcal{L}_{grad}
    &= \sum_{\mathcal{M} \in \Omega_h}\sum_{\br \in \mathcal{M}} \parallel \nabla ( D_{dpt}(\br) - (\hat D(\br) \cdot s+\jx{q})) \parallel_1 ,
\end{aligned}
\end{equation}
where $s$ and \jx{$q$} represent the scale and shift computed by linear least squares to convert the patches to the same scale following~\cite{ranftl2020towards}. 
Conversely, low uncertainty regions $\Omega_l$ contain more view-consistent correspondences with reliable depth, where the edge-aware smooth loss is employed as a regularization term to refine the continuity of the rendered depth. 
\begin{equation}
\mathcal{L}_{reg} = \sum_{\mathcal{M}  \in \Omega_l}\sum_{\br \in \mathcal{M}} \text{exp}(-\beta \nabla D_{dpt}(\br))\parallel \nabla \hat D(\br) \parallel_1,
\end{equation}
where $\beta$ denotes a hyperparameter to control the smooth extent. The exponential term serves as a weight that decreases as the depth gradient from DPT increases to preserve the edges, thereby promoting smoothness and continuity in low-uncertainty regions.

Through this uncertainty-guided mono-depth distillation, our method exploit the monocular geometry priors in the scene and provides balanced supervision with the uncertainty map as the reference. Our method not only incorporates detailed geometry and scale cues from single-view depth prediction but also aligns its understanding with the global, scale-aware information from SfM. It allows our model to effectively deal with various situations in the scene, enhancing the stability and performance of depth rendering.

\subsubsection{Training Loss} The total training loss for UC-NeRF is formulated by:
\begin{equation}
    \mathcal{L} = \lambda_1\mathcal{L}_{rgb} + \lambda_2 \mathcal{L}_{con} + \lambda_3 \mathcal{L}_{scale} + \lambda_4 \mathcal{L}_{grad} + \lambda_5 \mathcal{L}_{reg} ,
\end{equation}
where $\lambda_1, \lambda_2, \lambda_3, \lambda_4, \lambda_5$ denote the loss weights. In practice, we set  $\lambda_1=10, \lambda_2=0.5, \lambda_3=0.5, \lambda_4=0.5, \lambda_5=0.05$.

\section{Experiments}
\begin{table*}[t]
\small
\caption{\textbf{Quantitative Comparison of Novel View Synthesis Performance on SCARED Dataset.} $\nP$ denotes for NeRF with pre-training methods. $\nP$ w/ ft. denotes for the Pre-trained model after fine-tuning. $\nG$ denotes for NeRF with geometry guidance.\label{table: scared_comparison}}
\vspace{-3pt}
\resizebox{\linewidth}{!}{%
\renewcommand{\arraystretch}{1.1}
\begin{tabular}{@{}llcccccccc@{}}
\toprule
\multicolumn{1}{c}{\multirow{1}{*}{\textbf{Settings}}}& \textbf{Methods}&\textbf{PSNR}$ \uparrow$ &\textbf{SSIM}$ \uparrow$ &  \textbf{LPIPS}$\downarrow$&\textbf{Abs Rel}$\downarrow$ & \textbf{Sq Rel}$\downarrow$ & \textbf{RMSE}$\downarrow$ & \textbf{RMSE log}$\downarrow$&\textbf{$\boldsymbol{\delta}<\text{1.25}\uparrow$}
\\\cmidrule(l){1-10}
\multicolumn{1}{c}{\multirow{1}{*}{-}} &NeRF\cite{martin2021nerf} & 24.27\scriptsize$\pm$1.95& 0.712\scriptsize$\pm$0.10& 0.196\scriptsize$\pm$0.07& 0.324\scriptsize$\pm$0.12&2.776\scriptsize$\pm$2.14& 6.118\scriptsize$\pm$2.10&0.453\scriptsize$\pm$0.14& 0.470\scriptsize$\pm$0.12\\
\multicolumn{1}{c}{\multirow{1}{*}{\jx{-}}} &\jx{ActiveNeRF\cite{pan2022activenerf}} & \jx{23.29\scriptsize$\pm$1.56}& \jx{0.696\scriptsize$\pm$0.11}& \jx{0.215\scriptsize$\pm$0.08}& \jx{0.326\scriptsize$\pm$0.13}&\jx{2.927\scriptsize$\pm$2.54}& \jx{6.274\scriptsize$\pm$2.52}&\jx{0.433\scriptsize$\pm$0.14}& \jx{0.465\scriptsize$\pm$0.12}\\
\cmidrule(l){1-10}
\multicolumn{1}{c}{\multirow{3}{*}{ $\nP$}} &PixelNeRF\cite{yu2021pixelnerf} &15.24\scriptsize$\pm$1.57&0.432\scriptsize$\pm$0.08&0.308\scriptsize$\pm$0.04&0.327\scriptsize$\pm$0.08&2.371\scriptsize$\pm$1.43&5.794\scriptsize$\pm$1.83&0.363\scriptsize$\pm$0.06&0.531\scriptsize$\pm$0.09\\
 \multicolumn{1}{c}{} & MVSNeRF\cite{chen2021mvsnerf}  &17.35\scriptsize$\pm$1.63&0.582\scriptsize$\pm$0.07&0.408\scriptsize$\pm$0.04&0.266\scriptsize$\pm$0.05&2.020\scriptsize$\pm$1.01&5.401\scriptsize$\pm$1.87&0.318\scriptsize$\pm$0.07&0.536\scriptsize$\pm$0.08\\
\multicolumn{1}{c}{} &GeoNeRF\cite{johari2022geonerf} & 23.50\scriptsize$\pm$1.30&0.792\scriptsize$\pm$0.07& 0.152\scriptsize$\pm$0.05&\cellcolor{third}0.096\scriptsize$\pm$0.05&\cellcolor{third}0.196\scriptsize$\pm$0.13& \cellcolor{third}1.628\scriptsize$\pm$0.83& \cellcolor{third}0.093\scriptsize$\pm$0.04& 0.951\scriptsize$\pm$0.04\\\cmidrule(l){1-10}
\multicolumn{1}{c}{\multirow{3}{*}{ $\nP$ w/ ft.}} & PixelNeRF\cite{yu2021pixelnerf}  &21.45\scriptsize$\pm$1.87&0.673\scriptsize$\pm$0.09&0.288\scriptsize$\pm$0.08&0.271\scriptsize$\pm$0.11&1.479\scriptsize$\pm$1.24&4.890\scriptsize$\pm$1.63&0.301\scriptsize$\pm$0.08&0.817\scriptsize$\pm$0.05\\
   \multicolumn{1}{c}{}  &MVSNeRF\cite{chen2021mvsnerf} &23.09\scriptsize$\pm$1.69&0.751\scriptsize$\pm$0.07&0.248\scriptsize$\pm$0.06&0.246\scriptsize$\pm$0.04&1.357\scriptsize$\pm$1.12&4.021\scriptsize$\pm$1.39&0.280\scriptsize$\pm$0.06&0.854\scriptsize$\pm$0.05\\
 \multicolumn{1}{c}{} &GeoNeRF\cite{johari2022geonerf} &24.59\scriptsize$\pm$1.48&\cellcolor{second}0.825\scriptsize$\pm$0.10&\cellcolor{second}0.126\scriptsize$\pm$0.09&0.119\scriptsize$\pm$0.10&0.513\scriptsize$\pm$0.15&2.747\scriptsize$\pm$1.17&0.155\scriptsize$\pm$0.05&0.889\scriptsize$\pm$0.04\\
\cmidrule(l){1-10}
 \multicolumn{1}{c}{\multirow{6}{*}{$\nG$}} &NerfingMVS\cite{wei2021nerfingmvs}&22.70\scriptsize$\pm$1.76&0.598\scriptsize$\pm$0.05&0.285\scriptsize$\pm$0.05&0.228\scriptsize$\pm$0.11&1.275\scriptsize$\pm$1.61&4.251\scriptsize$\pm$2.06&0.290\scriptsize$\pm$0.08&0.598\scriptsize$\pm$0.14\\
 \multicolumn{1}{c}{} &RegNeRF\cite{niemeyer2022regnerf} & \cellcolor{third}25.18\scriptsize$\pm$1.41& 0.746\scriptsize$\pm$0.11&0.152\scriptsize$\pm$0.07&0.226\scriptsize$\pm$0.06&1.923\scriptsize$\pm$2.41&4.287\scriptsize$\pm$2.44& 0.276\scriptsize$\pm$0.06&\cellcolor{third}0.593\scriptsize$\pm$0.15\\ 
  \multicolumn{1}{c}{} &MonoSDF\cite{yu2022monosdf}&18.49\scriptsize$\pm$0.49&0.471\scriptsize$\pm$0.06&0.382\scriptsize$\pm$0.02&0.246\scriptsize$\pm$0.11&1.677\scriptsize$\pm$1.40&4.257\scriptsize$\pm$1.98&0.256\scriptsize$\pm$0.09&0.628\scriptsize$\pm$0.15\\ 
\multicolumn{1}{c}{} &DS-NeRF\cite{deng2022depth} & 24.93\scriptsize$\pm$1.55 &0.714\scriptsize$\pm$0.10&0.227\scriptsize$\pm$0.08&0.329\scriptsize$\pm$0.15&3.396\scriptsize$\pm$3.79&6.506\scriptsize$\pm$3.23&0.375\scriptsize$\pm$0.12&0.476\scriptsize$\pm$0.12\\

\multicolumn{1}{c}{} &Dense Prior\cite{roessle2022dense} &23.43\scriptsize$\pm$1.42&\cellcolor{third}0.816\scriptsize$\pm$0.05&\cellcolor{third}0.147\scriptsize$\pm$0.04&\cellcolor{second}0.088\scriptsize$\pm$0.03& \cellcolor{second}0.149\scriptsize$\pm$0.10&\cellcolor{second}1.458\scriptsize$\pm$0.66&\cellcolor{second}0.087\scriptsize$\pm$0.03& \cellcolor{second}0.932\scriptsize$\pm$0.04\\
\multicolumn{1}{c}{} &SparseNeRF\cite{wang2022sparsenerf} &\cellcolor{second}25.92\scriptsize$\pm$2.06&0.776\scriptsize$\pm$0.10&0.150\scriptsize$\pm$0.06&0.222\scriptsize$\pm$0.02&1.100\scriptsize$\pm$0.35&4.095\scriptsize$\pm$1.20&0.239\scriptsize$\pm$0.03&0.589\scriptsize$\pm$0.04\\
\multicolumn{1}{c}{} &\textbf{Ours} &\cellcolor{best}26.40\scriptsize$\pm$1.39&\cellcolor{best}0.855\scriptsize$\pm$0.05&\cellcolor{best}0.107\scriptsize$\pm$0.03&\cellcolor{best}0.053\scriptsize$\pm$0.03&\cellcolor{best}0.121\scriptsize$\pm$0.03&\cellcolor{best}1.304\scriptsize$\pm$0.68&\cellcolor{best}0.074\scriptsize$\pm$0.03&\cellcolor{best}0.965\scriptsize$\pm$0.04\\ \hline 
\end{tabular}

}
\vspace{-10pt}
\end{table*}
\begin{table*}[t]
\small
\centering
\caption{\textbf{Quantitative Comparison of Novel View Synthesis Performance on Hamlyn Dataset.} $\nP$ denotes for NeRF with pre-training methods. $\nP$ w/ ft. denotes for the Pre-trained model after fine-tuning. $\nG$ denotes for NeRF with geometry guidance.\label{table: hamlyn_comparison}}
\vspace{-3pt}
\resizebox{\linewidth}{!}{%
\renewcommand{\arraystretch}{1.1}
\begin{tabular}{@{}llcccccccc@{}}
\toprule
\multicolumn{1}{c}{\multirow{1}{*}{\textbf{Settings}}}&\textbf{Methods}&\textbf{PSNR}$ \uparrow$ &\textbf{SSIM}$ \uparrow$ &  \textbf{LPIPS}$\downarrow$&\textbf{Abs Rel}$\downarrow$ & \textbf{Sq Rel}$\downarrow$ & \textbf{RMSE}$\downarrow$ & \textbf{RMSE log}$\downarrow$&\textbf{$\boldsymbol{\delta}<\text{1.25}\uparrow$}
\\\cmidrule(l){1-10}
\multicolumn{1}{c}{\multirow{1}{*}{-}} &NeRF~\cite{martin2021nerf}&23.86\scriptsize$\pm$4.71&0.716\scriptsize$\pm$0.15&0.318\scriptsize$\pm$0.19&0.688\scriptsize$\pm$0.19&36.64\scriptsize$\pm$15.2&26.09\scriptsize$\pm$6.18&0.596\scriptsize$\pm$0.11&0.486\scriptsize$\pm$0.08\\
\multicolumn{1}{c}{\multirow{1}{*}{\jx{-}}} &\jx{ActiveNeRF\cite{pan2022activenerf}} & \jx{18.81\scriptsize$\pm$3.56}& \jx{0.536\scriptsize$\pm$0.14}& \jx{0.488\scriptsize$\pm$0.12}& \jx{0.770\scriptsize$\pm$0.18}&\jx{41.72\scriptsize$\pm$16.4}& \jx{32.94\scriptsize$\pm$6.15}&\jx{0.917\scriptsize$\pm$0.25}& \jx{0.327\scriptsize$\pm$0.07}\\
\cmidrule(l){1-10}
\multicolumn{1}{c}{\multirow{3}{*}{ $\nP$}} &PixelNeRF\cite{yu2021pixelnerf} & 14.21\scriptsize$\pm$3.47&0.621\scriptsize$\pm$0.09&0.524\scriptsize$\pm$0.09&0.690\scriptsize$\pm$0.15&34.06\scriptsize$\pm$8.97&25.77\scriptsize$\pm$4.16&0.547\scriptsize$\pm$0.11&0.439\scriptsize$\pm$0.09\\
 \multicolumn{1}{c}{} & MVSNeRF\cite{chen2021mvsnerf}&16.11\scriptsize$\pm$2.04&0.642\scriptsize$\pm$0.09&0.513\scriptsize$\pm$0.14&0.681\scriptsize$\pm$0.13&35.50\scriptsize$\pm$7.57&24.34\scriptsize$\pm$3.28&0.539\scriptsize$\pm$0.08&0.450\scriptsize$\pm$0.14\\
\multicolumn{1}{c}{} &GeoNeRF\cite{johari2022geonerf} &23.31\scriptsize$\pm$3.09&\cellcolor{third}0.802\scriptsize$\pm$0.09&\cellcolor{second}0.241\scriptsize$\pm$0.07&\cellcolor{second}0.596\scriptsize$\pm$0.18&{30.92\scriptsize$\pm$11.9}&\cellcolor{second}{17.84\scriptsize$\pm$6.54}&{0.498\scriptsize$\pm$0.13}&\cellcolor{second}0.682\scriptsize$\pm$0.26\\ \cmidrule(l){1-10}
\multicolumn{1}{c}{\multirow{3}{*}{$\nP$ w/ ft.}} & PixelNeRF\cite{yu2021pixelnerf}&21.45\scriptsize$\pm$3.74&0.659\scriptsize$\pm$0.12&0.470\scriptsize$\pm$0.10&0.663\scriptsize$\pm$2.14&32.91\scriptsize$\pm$3.30&22.37\scriptsize$\pm$3.02&0.524\scriptsize$\pm$0.07& 0.510\scriptsize$\pm$0.06\\
   \multicolumn{1}{c}{}  &MVSNeRF\cite{chen2021mvsnerf} &23.19\scriptsize$\pm$3.49&0.792\scriptsize$\pm$0.06&0.310\scriptsize$\pm$0.06&0.622\scriptsize$\pm$0.11&\cellcolor{third}29.42\scriptsize$\pm$7.89&19.86\scriptsize$\pm$1.87&0.501\scriptsize$\pm$0.05&0.541\scriptsize$\pm$0.03\\
 \multicolumn{1}{c}{} &GeoNeRF\cite{johari2022geonerf}&23.53\scriptsize$\pm$3.34& 0.783\scriptsize$\pm$0.11&0.268\scriptsize$\pm$0.11&0.621\scriptsize$\pm$0.16&31.12\scriptsize$\pm$12.1&19.70\scriptsize$\pm$5.76&0.545\scriptsize$\pm$0.12&0.617\scriptsize$\pm$0.22\\
\cmidrule(l){1-10}
 \multicolumn{1}{c}{\multirow{6}{*}{$\nG$}} &NerfingMVS\cite{wei2021nerfingmvs}&19.36\scriptsize$\pm$3.78&0.643\scriptsize$\pm$0.19&0.345\scriptsize$\pm$0.19&0.792\scriptsize$\pm$0.18&32.71\scriptsize$\pm$11.7&20.31\scriptsize$\pm$5.13&0.510\scriptsize$\pm$0.39&0.629\scriptsize$\pm$0.28\\
 \multicolumn{1}{c}{} &RegNeRF\cite{niemeyer2022regnerf} & \cellcolor{third}26.38\scriptsize$\pm$4.33&0.780\scriptsize$\pm$0.15&0.274\scriptsize$\pm$0.21&0.703\scriptsize$\pm$0.21&38.56\scriptsize$\pm$13.9&24.45\scriptsize$\pm$6.35& 0.594\scriptsize$\pm$0.12&0.515\scriptsize$\pm$0.21\\ 
  \multicolumn{1}{c}{} &MonoSDF\cite{yu2022monosdf} & 20.20\scriptsize$\pm$3.69&0.701\scriptsize$\pm$0.08&0.473\scriptsize$\pm$0.07&0.628\scriptsize$\pm$0.20&\cellcolor{second}28.77\scriptsize$\pm$15.0&19.48\scriptsize$\pm$7.37&\cellcolor{second}0.472\scriptsize$\pm$0.10& 0.506\scriptsize$\pm$0.17\\ 
\multicolumn{1}{c}{}&DS-NeRF\cite{deng2022depth}&24.12\scriptsize$\pm$7.82&0.687\scriptsize$\pm$0.24&0.351\scriptsize$\pm$0.29&0.692\scriptsize$\pm$0.11&35.43\scriptsize$\pm$7.47&23.86\scriptsize$\pm$4.07&0.741\scriptsize$\pm$0.49&0.463\scriptsize$\pm$0.14\\
\multicolumn{1}{c}{} &Dense Prior\cite{roessle2022dense} & 23.79\scriptsize$\pm$4.52&0.710\scriptsize$\pm$0.18&\cellcolor{third}{0.257\scriptsize$\pm$0.21}&0.616\scriptsize$\pm$0.17&30.45\scriptsize$\pm$11.2&19.43\scriptsize$\pm$4.61&0.502\scriptsize$\pm$0.29&\cellcolor{third}{0.636\scriptsize$\pm$0.20}\\
\multicolumn{1}{c}{} &SparseNeRF\cite{wang2022sparsenerf} &\cellcolor{second}26.72\scriptsize$\pm$3.64&\cellcolor{second}0.812\scriptsize$\pm$0.13&0.293\scriptsize$\pm$0.22&\cellcolor{third}0.610\scriptsize$\pm$0.18&30.61\scriptsize$\pm$11.8&\cellcolor{third}18.50\scriptsize$\pm$4.62&\cellcolor{third}0.485\scriptsize$\pm$0.08&0.585\scriptsize$\pm$0.22\\
\multicolumn{1}{c}{} &\textbf{Ours} & \cellcolor{best}26.87\scriptsize$\pm$1.89&\cellcolor{best}0.819\scriptsize$\pm$0.07&\cellcolor{best}0.215\scriptsize$\pm$0.06&\cellcolor{best}0.530\scriptsize$\pm$0.15&\cellcolor{best}26.37\scriptsize$\pm$9.52&\cellcolor{best}16.15\scriptsize$\pm$3.59&\cellcolor{best}0.450\scriptsize$\pm$0.07&\cellcolor{best}0.741\scriptsize$\pm$0.19\\ \hline 
\end{tabular}

}
\vspace{-10pt}
\end{table*}
\subsection{Implementation Details}
\subsubsection{Experimental Setup}
The code of our method is based on PyTorch, running on NVIDIA GeForce RTX 3090 GPU. We process rays for scale distillation in a batch size of 1024, and patches for guided patch sampling with size of $6\times6$ in batch size of 50. We use Adam Optimizer for our network, with a learning rate of $6 \times 10 ^{-4}$ with a cosine decay scheduler. For the sampling points on a ray, we adopt 90 points during training and inference for efficiency. For the consistency learner, we utilize a pre-trained Cas-MVSNet model~\cite{gu2020cascade}, with the depth hypothesis planes decreasing as (48, 32, 8) in three stages. All the experiments and comparisons use the image size of $320 \times 256$. To compare with NeRF-based methods which require no pre-training, we train the model from scratch. For NeRF with pre-training methods, we utilize their released pre-trained model and train the fine-tuned model on each scene.
\subsubsection{Datasets} 
We train our method on the SCARED Dataset~\cite{allan2021stereo} and Hamlyn Dataset~\cite{mountney2010three, stoyanov2010real, pratt2010dynamic}, which contain challenging endoscopic scenes with weak textures, reflections, and occlusions. \jxg{Following the preprocessing in~\cite{roessle2022dense}, after filtering out the frames with motion blur or flaws, we obtained 9 and 6 scenes from the SCARED and Hamlyn datasets respectively. Specifically, we first use a sliding window approach and extract the sharpest image from every N frame. Images with severe occlusions are also removed to avoid introducing noise into the training process. 
Next, we downsample the dataset to enhance computational efficiency and avoid redundancy from very similar frames.
Specifically, we manually select frames to ensure each frame provides unique coverage of the surgical scene, with an overlap of approximately 60-80 $\%$ between consecutive frames following \cite{roessle2022dense}. 
This step leads to reduced data size, maintaining a balance between comprehensive scene coverage and training efficiency.
}

For Hamlyn Dataset, we follow the volumetric reconstruction part of Endo-Depth-and-Motion~\cite{recasens2021endo} to specifically focus on static scenes with larger camera translation to validate our method. 
Note that the collected scenes only contain monocular images, taken from the left camera view for both SCARED and Hamlyn Dataset. We use COLMAP~\cite{schoenberger2016sfm} to preprocess the datasets to get the camera poses and sparse pointclouds following \cite{mildenhall2021nerf, deng2022depth}. We obtain the monocular depth from DPT~\cite{ranftl2021vision} as priors to guide our UC-NeRF. Similar to the data split in \cite{mildenhall2021nerf, niemeyer2022regnerf, wei2021nerfingmvs}, we collect 20 images covering a local area for each scene and hold out 1/2 of these views for testing and the remaining for training.
During training, we take 7 closest neighboring views as source views for the consistency learner to extract features and 3D neural volumes.
\begin{figure*}[t]
\centering
\includegraphics[width=0.95\linewidth]{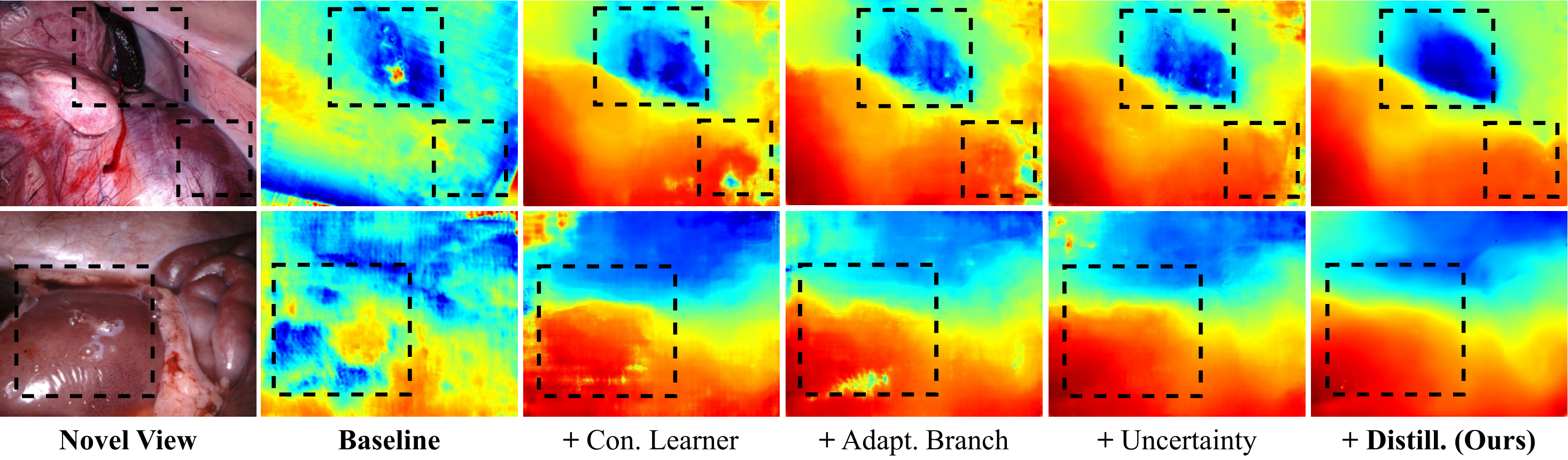}
\vspace{-4pt}
\caption{\jxg{\textbf{Visualization of the ablation study on different components.} 
Using all the components achieves the best depth estimation.}
}
\label{fig: abl_compo}
\vspace{-10pt}
\end{figure*}

\subsubsection{Metrics}
We report the quantitative comparisons of appearance and geometry to show the synthesis performance, including the mean and standard deviation across the test novel views. We further highlight the table cell to show \colorbox{best}{best}, \colorbox{second}{second best}, and \colorbox{third}{third best}. We adopt the PSNR, SSIM~\cite{wang2004image} and LPIPS~\cite{zhang2018unreasonable} for assessing image synthesis quality. To evaluate the depth results, we scale the predicted depth maps with median scaling~\cite{zhou2017unsupervised}.
Following~\cite{wei2021nerfingmvs}, we adopt the error and accuracy metrics for depth evaluation, including Abs Rel, Sq Rel, RMSE in mm, RMSE log in log mm and $\delta < t$ in \%.

\subsection{Comparison Study}
We report the quantitative results in Tab.~\ref{table: scared_comparison} and Tab.~\ref{table: hamlyn_comparison}, and qualitative comparison in Fig.~\ref{fig: qualitative}. \jx{We first compare with NeRF~\cite{mildenhall2021nerf} and ActiveNeRF~\cite{pan2022activenerf} to validate the performance of our multi-view uncertainty estimation in dealing with endoscopic sparse views. Our method achieves consistent improvement on both color and geometry rendering compared to ActiveNeRF~\cite{pan2022activenerf} that incorporates the uncertainty by modeling the rendered radiance value as a Gaussian distribution.}

\subsubsection{NeRF w/ Pre-training}We compare both the pre-trained generalizable model and fine-tuned model on the SCARED dataset and the Hamlyn dataset. PixelNeRF~\cite{yu2021pixelnerf}, MVSNeRF~\cite{chen2021mvsnerf}, GeoNeRF~\cite{johari2022geonerf} have trained generalizable model on large datasets (DTU~\cite{jensen2014large}, LLFF~\cite{mildenhall2019local}, IBRNet~\cite{wang2021ibrnet}) with large cost in time and computation. However, due to the domain gap between general scene and surgical scene, directly adapting the pre-trained generalizable NeRF model to the surgical scene shows unsatisfying results. As shown in Fig.~\ref{fig: qualitative}, MVSNeRF~\cite{chen2021mvsnerf} and GeoNeRF~\cite{johari2022geonerf} produce blurs and flaws in the rendered color, and depth with floating artifacts and noises on the surgical tissue. After fine-tuning each scene, while the RGB performance has been improved, the geometry performance degenerates due to shape-radiance ambiguity, especially in the region with changing illumination and reflective surface. 

\begin{figure*}[t]
\centering
\includegraphics[width=0.95\linewidth]{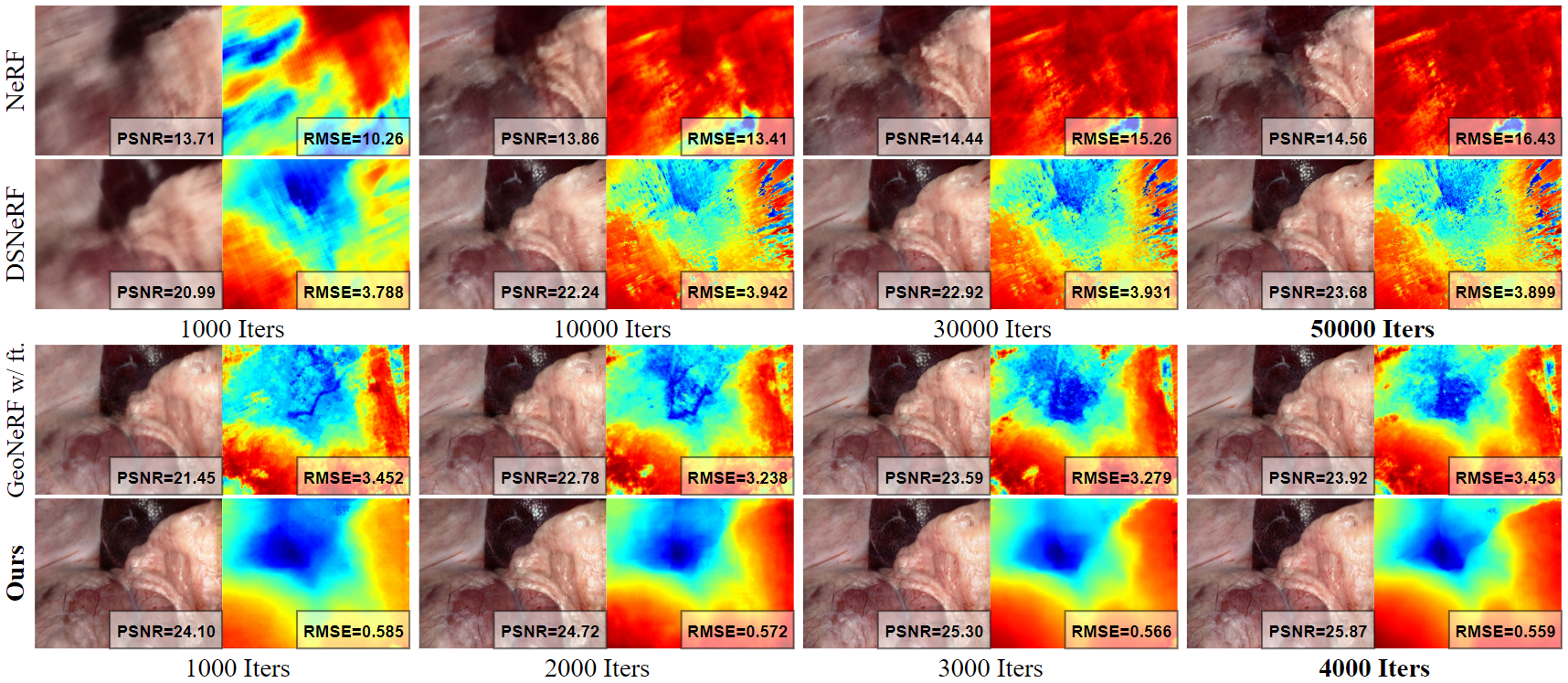}
\vspace{-4pt}
\caption{\textbf{Visualization for the ablation study on training efficiency}. Our UC-NeRF achieves the most efficient and effective training.}
\label{fig: efficiency}
\vspace{-5pt}
\end{figure*}

\begin{table*}[h]
\begin{minipage}{.49\textwidth}
    \caption{\jx{\textbf{Computation cost of different methods} on the SCARED Dataset~\cite{allan2021stereo}. 
    We compare the time and GPU memory required for training, and render time during inference.}
    }
    \centering
    \vspace{-1mm}
    \renewcommand\theadfont{\fsize}
\resizebox{\linewidth}{!}{%
\renewcommand{\arraystretch}{1.05}
\setlength{\tabcolsep}{1pt}
    \begin{tabular}{@{}lccc@{}}
    \toprule
    \multirow{2}{*}{\textbf{Methods}} & \multicolumn{2}{c}{\textbf{Training}}& \textbf{Inference} \\ \cmidrule(lr){2-4}
    \multicolumn{1}{c}{} & \textbf{GPU Hour $\downarrow$} & \textbf{GPU Mem. $\downarrow$} &\textbf{Render Time $\downarrow$} \\
    \hline
    NeRF~\cite{mildenhall2021nerf} & 34.48h& 3.705G & 1.855s  \\
    ActiveNeRF~\cite{pan2022activenerf} & 35.08h & 4.365G &  2.655s \\
    RegNeRF~\cite{niemeyer2022regnerf}  &25.84h&21.94G & 17.75s \\
    DSNeRF~\cite{deng2022depth}  & 362.5h & 10.21G& 1.846s \\
    Dense Prior~\cite{roessle2022dense} & 35.71h& 5.681G &1.834s\\
    SparseNeRF~\cite{wang2022sparsenerf}  &49.52h &21.94G  & 19.24s\\
    GeoNeRF w/ ft.~\cite{johari2022geonerf}$^\dagger$  & 2.529h& 10.95G  & 7.796s\\
    \textbf{UC-NeRF (Ours)} & 2.201h& 7.638G & 0.828s  \\
    \hline
    \end{tabular}
    }
    \vspace{-2mm}
    \begin{flushleft}
\footnotesize $^\dagger$Pretrained on DTU~\cite{jensen2014large}, LLFF~\cite{mildenhall2019local}, IBRNet~\cite{wang2021ibrnet}.
\end{flushleft}
    \label{tab:efficiency_tab}	
\end{minipage}
\hfill
	\begin{minipage}{.49\textwidth}
	\centering
 \begin{figure}[H]
     \centering
     \includegraphics[width=\linewidth]{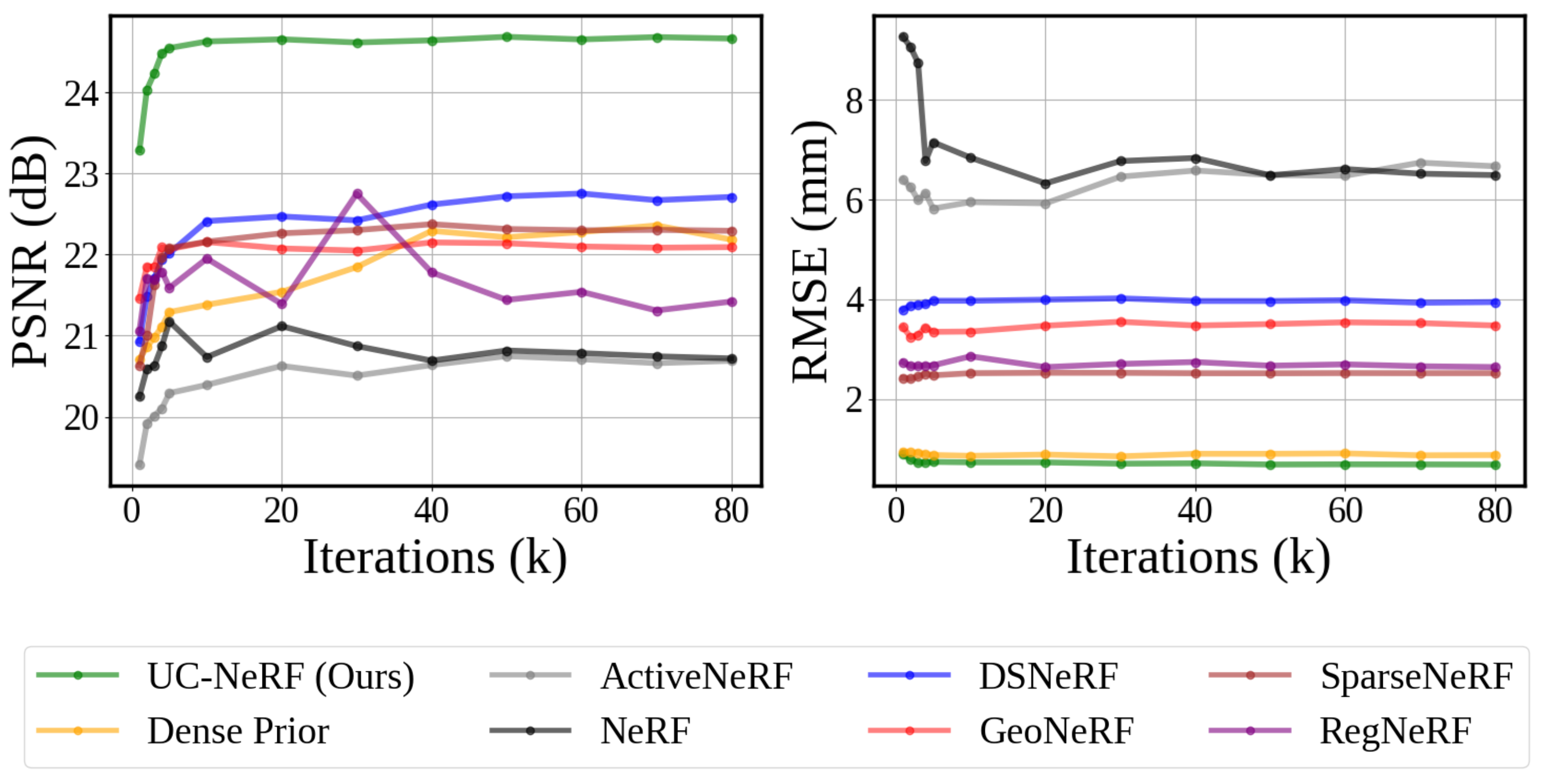}
     \caption{\jx{\textbf{Training efficiency.} We visualize the charts of PSNR-iteration and RMSE-iteration for training ``scene 09" in the SCARED Dataset~\cite{allan2021stereo}. }}
     \label{fig:efficiency_curve}
 \end{figure}
	\end{minipage}%
    \vspace{-6mm}
\end{table*}
\subsubsection{NeRF w/ Geometry Guidance}
 To enhance the geometry learning of the radiance field, NerfingMVS~\cite{wei2021nerfingmvs}, DS-NeRF ~\cite{deng2022depth}, Dense Depth Priors~\cite{roessle2022dense} take advantage of the sparse depth priors from COLMAP during the pre-processing stage, as constraints for ray sampling range or supervision to rendered depth. However, as presented in Fig.~\ref{fig: qualitative}, they fail to tackle the severe photometric inconsistencies under sparse surgical views, causing deficiencies in rendered color and depth. 
 
While leveraging multiple monocular geometric priors, MonoSDF~\cite{yu2022monosdf} exhibits blurred rendered color, lack of detail, and overly smooth surfaces in depth. 
RegNeRF~\cite{niemeyer2022regnerf} addresses input sparse views by introducing a patch-based regularizer that improves performance. Although its color reconstruction outperforms other baselines, the degeneration in depth indicates suffering from shape-radiance ambiguity. We further compare our method with the recent SparseNeRF~\cite{wang2022sparsenerf} which regularizes the rendered depth with depth ranking and continuity constraints. Despite its improvements in color rendering, it shows limited depth promotion due to its unified optimization for regions differing in uncertainty.
Our UC-NeRF demonstrates robustness to outliers with uncertainty guidance, resulting in accurate depth while preserving view-dependent effects. It achieves consistent improvement in both color and depth rendering results, with higher mean and lower standard deviation compared to baselines.
\subsection{Ablation Study}
\subsubsection{Different Components of UC-NeRF}
\jxg{We conduct an ablation study by gradually adding more components to analyze the effectiveness of different components in our UC-NeRF}, presenting the qualitative results in Fig. 6 and quantitative comparisons in Tab. IV. Specifically, we utilize MVSNeRF~\cite{chen2021mvsnerf} as the baseline method for comparison. 
\jxg{The notation ``+ Con. Learner" indicates the addition of our proposed consistency learner. 
The notation ``+ Adapt. Branch" refers to further including the adaptive branch and simply summing up both branches' prediction.
The notation ``+ Uncertainty" denotes the proposed uncertainty-aware adaptation used to fuse the two branches (i.e. weighted sum). 
Finally, the notation ``+ Distill." denotes the inclusion of distillation from monocular geometry priors.} Adding the consistency learner helps extract view-consistent correspondences from sparse views with guidance from SfM, facilitating both the geometry and color rendering. Then, the adaptive branch enhances NeRF with the high-level feature, however, causing degeneration in geometry due to the photometric inconsistency (e.g. the hole in the reflective surface in example 2). To address this problem, the uncertainty-aware adaptation merges the output from the base-adaptive branch using the uncertainty map, adaptively handling the photometric inconsistency. Finally, the distillation from monocular geometry priors further enhances the performance.

\subsubsection{Effect of deep feature} We validate the effectiveness of the high-level image features in recovering the view-dependent effect at the adaptive branch. As shown in Fig.~\ref{fig: branch_vis}, integrated with the deep features, the adaptive branch is capable of capturing high-frequency details and view-dependent effects, \eg reflective surfaces, tiny veins, and blood. 
\subsubsection{Efficiency} \jx{To evaluate the efficiency of our UC-NeRF, we visualize the average metrics changing with iterations of ``scene 09'' in the SCARED Dataset in Fig.~\ref{fig:efficiency_curve} and present the qualitative result in Fig.~\ref{fig: efficiency}. 
It shows that both the PSNR curve and RMSE curve of our method converge faster in the beginning iterations, while other methods converge slowly and yield low performance. }As depicted in Fig.~\ref{fig: efficiency}, only after 1000 iterations of training, our method has achieved PSNR 24.10 and RMSE 0.586, which outperforms NeRF~\cite{mildenhall2021nerf} and DSNeRF~\cite{deng2022depth} in both geometry and color rendering after 50000 iters training. Even after generic training on multiple large datasets, GeoNeRF\cite{johari2022geonerf} still yields unsatisfying performance in surgical datasets which have large domain gap to general scenes. \jx{To fully compare the training and inference efficiency, we report the time (GPU Hour) and GPU memory for training, and render time during inference in Tab.~\ref{tab:efficiency_tab}. Thanks to our generalizable training across multiple surgical scenes, our UC-NeRF demonstrates superior efficiency in training time. With few hypothesis planes in the consistency learner and a reduction of sampled points during NeRF rendering, the inference time surpasses other state-of-the-art baselines. }

\begin{table*}[t]
\small
\caption{\jxg{\textbf{Ablation study on different training data size}.\label{table: diff-sample-ratio}}}
\renewcommand\theadfont{\fsize}
\resizebox{\linewidth}{!}{%
\renewcommand{\arraystretch}{1.1}
\begin{tabular}{@{}lcccccccccc@{}}
\toprule
\textbf{Training Size} &\textbf{$\text{PSNR}\uparrow$}& \textbf{$\text{SSIM}\uparrow$}  & \textbf{$\text{LPIPS}\downarrow$} & \textbf{Abs Rel}$\downarrow$ & \textbf{Sq Rel}$\downarrow$ & \textbf{RMSE}$\downarrow$ & \textbf{RMSE log}$\downarrow$&\textbf{$\boldsymbol{\delta}<\text{1.25}\uparrow$}&\textbf{GPU Hour}$\downarrow$
\\\midrule %
50 & \cellcolor{best}26.73\scriptsize$\pm$1.83& \cellcolor{best}0.860$\pm$0.06&\cellcolor{third}0.109$\pm$0.04&\cellcolor{second}0.055$\pm$0.04 &\cellcolor{third} 0.128$\pm$0.04& \cellcolor{third}1.327$\pm$0.63 &\cellcolor{third}0.082$\pm$0.05&\cellcolor{second}0.967$\pm$0.05 & \cellcolor{third}10.83h\\
30 & \cellcolor{second}26.49\scriptsize$\pm$1.44& \cellcolor{second}0.857$\pm$0.06&\cellcolor{second}0.107$\pm$0.05&\cellcolor{third}0.056$\pm$0.03 & \cellcolor{second}0.126$\pm$0.05& \cellcolor{second}1.317$\pm$0.55 &\cellcolor{second}0.077$\pm$0.03&\cellcolor{best}0.965$\pm$0.04 & \cellcolor{second}6.423h\\
\textbf{10 (Ours)} & \cellcolor{third}26.40\scriptsize$\pm$1.39& \cellcolor{third}0.855$\pm$0.05&\cellcolor{best} 0.107$\pm$0.03& \cellcolor{best}0.053$\pm$0.03& \cellcolor{best}0.121$\pm$0.04 & \cellcolor{best}1.304$\pm$0.68 &\cellcolor{best}0.074$\pm$0.03&\cellcolor{best}0.965$\pm$0.04&\cellcolor{best} 2.201h\\
\hline 
\end{tabular}

}
\vspace{-12pt}
\end{table*}
\begin{table}
\begin{minipage}{.5\textwidth}
\small
\caption{\textbf{\jxg{Ablation study on the influence of different components}}. \label{table: ablation}}
\renewcommand\theadfont{\fsize}
\resizebox{\linewidth}{!}{%
\renewcommand{\arraystretch}{1.1}
\begin{tabular}{@{}lccccc@{}}
\toprule
\textbf{Components} &\textbf{$\text{PSNR}\uparrow$}&  \textbf{$\text{SSIM}\uparrow$} & \textbf{$\text{LPIPS}\downarrow$} & \textbf{$\text{RMSE}\downarrow$} 
\\ \midrule
Baseline  &22.21\scriptsize$\pm$1.68&0.652\scriptsize$\pm$0.08&0.250\scriptsize$\pm$0.07&5.069\scriptsize$\pm$1.88\\
 \jxg{+ Con. Learner}&25.18\scriptsize$\pm$1.60&0.808\scriptsize$\pm$0.07&0.163\scriptsize$\pm$0.07&\cellcolor{third}1.875\scriptsize$\pm$0.82\\
 \jxg{+ Adapt. Branch}& \cellcolor{third}25.78\scriptsize$\pm$1.69& \cellcolor{third}0.830\scriptsize$\pm$0.06& \cellcolor{third}0.119\scriptsize$\pm$0.06&1.895\scriptsize$\pm$1.27\\
\jxg{+ Uncertainty}  & \cellcolor{second}26.12\scriptsize$\pm$1.45& \cellcolor{second}0.843\scriptsize$\pm$0.06& \cellcolor{second}0.115\scriptsize$\pm$0.03&\cellcolor{second}1.524\scriptsize$\pm$0.81\\
\textbf{\jxg{+ Distill. (Ours)}} & \cellcolor{best}26.40\scriptsize$\pm$1.39& \cellcolor{best}0.855\scriptsize$\pm$0.05& \cellcolor{best}0.107\scriptsize$\pm$0.03& \cellcolor{best}1.304\scriptsize$\pm$0.68\\
\hline 
\end{tabular}

}
\vspace{5pt}
\end{minipage}
\begin{minipage}{.5\textwidth}

\small
\caption{\textbf{Ablation study on the number of source views}.\label{table: view_num}}
\renewcommand\theadfont{\fsize}
\resizebox{\linewidth}{!}{%
\renewcommand{\arraystretch}{1.1}
\begin{tabular}{@{}lccccc@{}}
\toprule
\textbf{Number of Views} &\textbf{$\text{PSNR}\uparrow$}& \textbf{$\text{SSIM}\uparrow$}  & \textbf{$\text{LPIPS}\downarrow$} & \textbf{$\text{RMSE}\downarrow$} 
\\\midrule %
3 Views  &26.13\scriptsize$\pm$1.24&0.841\scriptsize$\pm$0.05& \cellcolor{third}0.116\scriptsize$\pm$0.04& 1.474\scriptsize$\pm$0.71\\
5 Views& \cellcolor{second}26.20\scriptsize$\pm$1.22& \cellcolor{third}0.845\scriptsize$\pm$0.05& \cellcolor{second}0.115\scriptsize$\pm$0.04& \cellcolor{third}1.393\scriptsize$\pm$0.64\\
\textbf{7 Views}& \cellcolor{best}{26.40}\scriptsize$\pm$1.39& \cellcolor{best}{0.855}\scriptsize$\pm$0.05&\cellcolor{best}{0.107}\scriptsize$\pm$0.03&\cellcolor{best}{1.304}\scriptsize$\pm$0.68\\
9 Views& \cellcolor{third}26.15\scriptsize$\pm$1.35& \cellcolor{second}0.847\scriptsize$\pm$0.06& 0.118\scriptsize$\pm$0.03& \cellcolor{second}1.321\scriptsize$\pm$0.71\\
\hline 
\end{tabular}

}
\vspace{-4pt}
\end{minipage}
\end{table}
\subsubsection{Robustness to Number of Source Views} 
\jx{We conduct an ablation study to validate our method's robustness to different input source views for the consistency learner.
During training, the input source views are selected as the $N$ closest views to the target view. 
As shown in Tab.~\ref{table: view_num}, our method is robust to the number of input views and achieves the best performance with 7 input views. 
The slight performance drop from using 7 views to 9 is probably because using too many source views may introduce excessive information into training, leading to information redundancy with distractions or noises for rendering the target view.}

\subsubsection{Robustness to Different Training Data Size} 
\jxg{We conduct an ablation study to validate the robustness to training data sizes (i.e. total number of training images).
Specifically, after filtering out the low-quality images, we evenly sample different numbers of images (i.e. 50, 30, 10) as training images and validate on the same test set. 
As shown in Tab.~\ref{table: diff-sample-ratio}, it is observed that our sparse set shows competitive performance with the dense training set while being much more efficient. With the training size increases, it is observed that the performance of rendered images becomes better, because the source views would contain more contextual information to the target view.}

\jxg{This experiment helps validate that our method can tackle surgical datasets of different sizes, showing robustness to the endoscopic sparse views and demonstrating the potential for real-world surgery where data acquisition opportunities are naturally restricted.}
\section{Discussion}
\begin{figure}[t]
\centering
\includegraphics[width=0.95\linewidth]{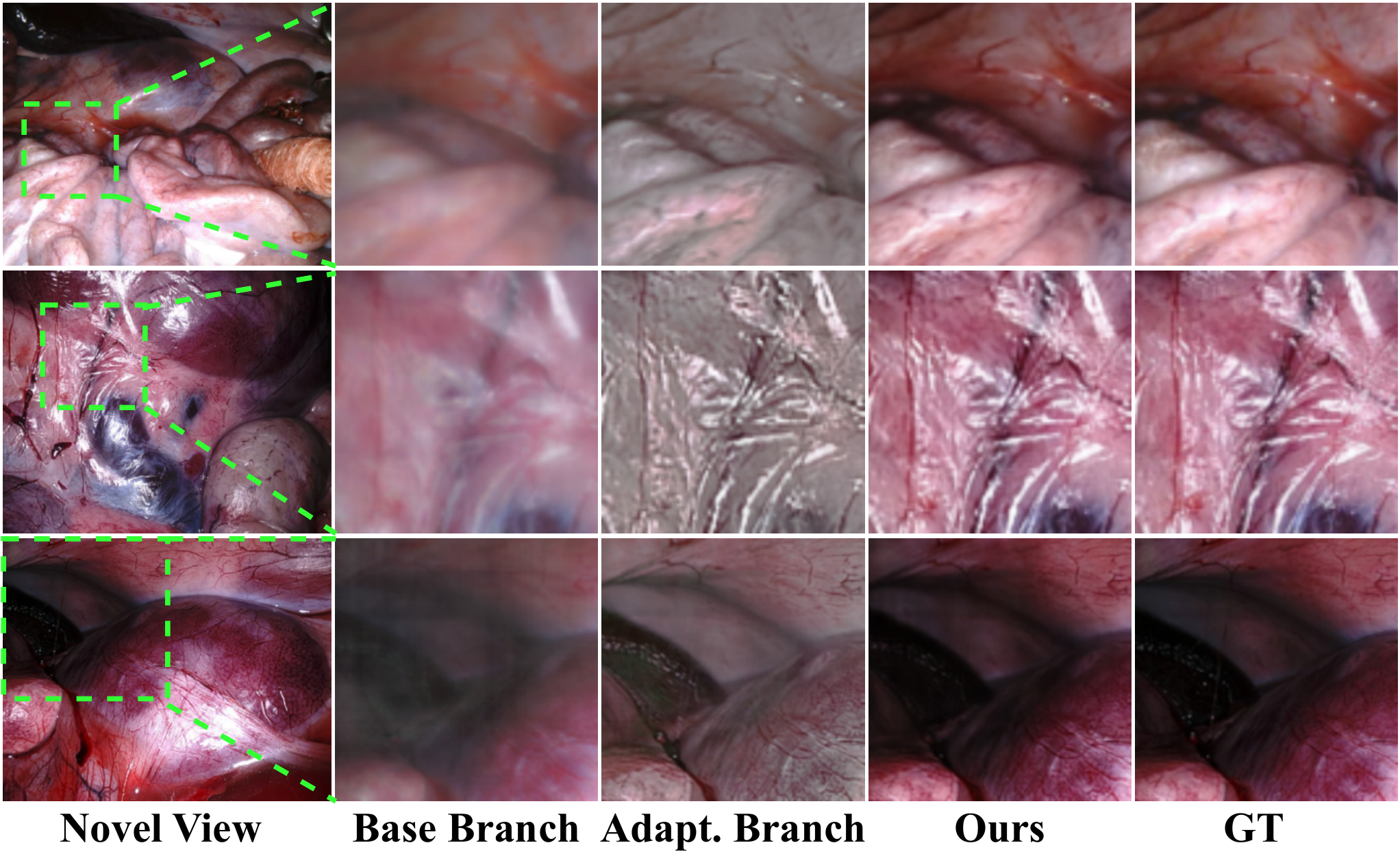}
\caption{\textbf{View synthesis decomposition in dual-branch NeRF}. We decompose the synthesized images in the base branch, adaptation branch, and the full model of UC-NeRF. It shows the ability of the high-level features to model view-dependent effects.}
\label{fig: branch_vis}
\vspace{-10pt}
\end{figure}

The objective of this paper is to improve NeRF's ability to handle the shape-ambiguity problem caused by challenges in surgical scenes, such as endoscopic sparse views and photometric inconsistencies. Our method is the first to address the challenging sparse view NeRF problem in surgical scenes. The proposed UC-NeRF is both effective and efficient in surgical novel view synthesis because of three main reasons. First, the view-specific uncertainty information is estimated through the consistency learner guided by the sparse SfM depth to measure the extent of photometric inconsistency across multi-view inputs (See Fig.~\ref{fig: uncertainty}). Moreover, our method is capable of generic training across multiple surgical scenes thanks to the conditional inputs from the consistency learner. Second, unlike other few-shot NeRF methods to inject the unified constraints, we take benefits of the learnt uncertainty information to explicitly model the regions in the designed dual-branch NeRF, \ie fusing the base and adaptive branch with different weight to generate view-dependent appearance and consistent geometry. This design is further demonstrated in Fig.~\ref{fig: branch_vis} to visualize the different modeling effects decomposed in the two branches. Third, we further introduce the distillation from monocular geometry priors to improve the accuracy and robustness of the rendered depth (see Fig.~\ref{fig: abl_compo}). Specifically, the scale distillation improves the scale consistency through the sparse SfM depth constraints. The uncertainty-guided mono-depth distillation employ the supervision and regularization adaptively in spatial following the uncertainty estimation. 

Despite the aforementioned strengths, our method is limited to static or semi-static surgical scene, which hinders its application to the highly dynamic surgical scenes. In the future work, we will try to integrate time dimension to enable the spatialtemporal NeRF, allowing both time-varying and view-free rendering for color and depth. More efficient neural representations are also considered to integrate with our method to improve the efficiency for real-time rendering, which benefits the downstream tasks like surgical navigation, 3D reconstruction, surgical skill learning, \etc.

\section{Conclusions}
In this paper, the UC-NeRF network addresses the difficulties posed by surgical sparse views and photometric inconsistencies, achieving robust and efficient novel view synthesis in minimally invasive surgery. By leveraging a consistent learner and an uncertainty map, UC-NeRF enhances geometric correspondence and significantly reduces shape-radiance ambiguity. The introduction of an uncertainty-aware conditional NeRF also refines the learning process of view-dependent appearances, ensuring a balance between geometric precision and photorealistic rendering. Our experimental evaluation shows that our method outperforms other few-shot NeRF methods in both efficiency and effectiveness.

\bibliographystyle{IEEEtran}
\bibliography{egbib}

\end{document}